\def\year{2016}\relax
\documentclass[letterpaper]{article}
\usepackage{aaai16}
\usepackage{times}
\usepackage{helvet}
\usepackage{courier}
\usepackage{floatrow}
\usepackage{graphicx}

\usepackage{amsmath, amssymb}
\newtheorem{theorem}{Theorem}
\newtheorem{proposition}{Proposition}
\newtheorem{lemma}{Lemma}
\newtheorem{definition}{Definition}

\def\myproof{1} 
\if\myproof1\nocopyright\fi

\begin{document}
\title{Near-Optimal Active Learning of Multi-Output Gaussian Processes}
\author{
Yehong Zhang$^{\dag}$ \and Trong Nghia Hoang$^{\ast}$ \and Kian Hsiang Low$^{\dag}$ \and Mohan Kankanhalli$^{\dag}$\\
Department of Computer Science$^{\dag}$, Interactive Digital Media Institute$^{\ast}$\\ 
National University of Singapore, Republic of Singapore\\
\{yehong, lowkh, mohan\}@comp.nus.edu.sg$^{\dag}$, idmhtn@nus.edu.sg$^{\ast}$ 
%Paper ID: \#2548\\
%Keywords: Active learning, Gaussian process, multi-output Gaussian process\\
%Topics: ML: Active learning, ML: Transfer, adaptation, multitask learning,\\
%CSAI: Modeling and prediction of dynamic and spatiotemporal phenomena and systems
}
\maketitle
\begin{abstract}
\begin{quote}
%AAAI creates proceedings, working notes, and technical reports directly from electronic source furnished by the authors. To ensure that all papers in the publication have a uniform appearance, authors must adhere to the following instructions. 
%This paper presents novel active learning strategies for multi-output Gaussian process (MOGP) model which can formally characterize the spatial correlation and cross-correlation structure for multiple correlated phenomena in an environment. Our algorithm, in contrast to existing works, selects both the most informative sampling locations and the type of measurements for minimizing the joint predictive uncertainty of only the target phenomenon. To make the algorithm scales well in the number of possible sampling locations and observations, we exploit the sparse structure of MOGP for deriving a novel active learning criterion and an efficient greedy algorithm with performance guarantee for the new criterion. 
This paper\if\myproof1\footnote{This paper is an extended version (with proofs) of \cite{YehongAAAI16}.}\fi addresses the problem of active learning of a \emph{multi-output Gaussian process} (MOGP) model representing multiple types of coexisting correlated environmental phenomena. In contrast to existing works, our active learning problem involves selecting not just the most informative sampling locations to be observed but also the types of measurements at each selected location for minimizing the predictive uncertainty (i.e., posterior joint entropy) of a target phenomenon of interest given a sampling budget.
Unfortunately, such an entropy criterion scales poorly in the numbers of candidate sampling locations and selected observations when optimized.
To resolve this issue, we first exploit a structure common to sparse MOGP models for deriving a novel active learning criterion. Then, we exploit a relaxed form of submodularity property of our new criterion for devising a polynomial-time approximation algorithm that guarantees a constant-factor approximation of that achieved by the optimal set of selected observations.
Empirical evaluation on real-world datasets shows that our proposed approach outperforms existing algorithms for active learning of MOGP and single-output GP models.
\end{quote}
\end{abstract}

\section{Introduction}
\label{sect:intro}
For many budget-constrained environmental sensing and monitoring applications in the real world, 
active learning/sensing is an attractive, frugal alternative to passive high-resolution (hence, prohibitively costly) sampling of the spatially varying target phenomenon of interest. Different from the latter, active learning aims to select and gather the \emph{most informative} observations for modeling and predicting the spatially varying phenomenon given some sampling budget constraints (e.g., quantity of deployed sensors, energy consumption, mission time). 

In practice, the target phenomenon often coexists and correlates well with some auxiliary type(s) of phenomena whose measurements
%is often associated with some highly correlated auxiliary measurement type(s) that 
may be more spatially correlated, less noisy (e.g., due to higher-quality sensors), and/or less tedious to sample (e.g., due to greater availability/quantity, higher sampling rate, and/or lower sampling cost of deployed sensors of these type(s)) and can consequently be exploited for improving its prediction.
For example, to monitor soil pollution by some heavy metal (e.g., Cadmium), its complex and time-consuming extraction from soil samples can be alleviated by supplementing its prediction with correlated auxiliary types of soil measurements (e.g., pH) that are easier to sample \cite{Goovaerts97b}.
%It is, however, easier to measure the pH level which is spatially correlated with the heavy metal concentration
Similarly, to monitor algal bloom in the coastal ocean, plankton abundance correlates well with auxiliary types of ocean measurements (e.g., chlorophyll a, temperature, and salinity)  that can be sampled more readily. 
% \cite{Apple08}\cite{Atkinson2000}
Other examples of real-world applications include remote sensing, traffic monitoring \cite{LowUAI12,LowRSS13,LowTASE15}, monitoring of groundwater
% \cite{Passarella2003} 
and indoor environmental quality \cite{Xu2014}, 
%\cite{Choi12}
and precision agriculture \cite{Webster01}, among others. 
%as well as that pertaining to the Web such as natural language processing \cite{Reichart2008} and recommender systems \cite{Zhao2013}, among others.  
All of the above practical examples motivate the need to design and develop an active learning algorithm that selects not just the \emph{most informative} sampling locations to be observed but also the types of measurements (i.e., target and/or auxiliary) at each selected location for minimizing the predictive uncertainty of unobserved areas of a target phenomenon given a sampling budget, which is the focus of our work here\footnote{Our work here differs from multivariate spatial sampling algorithms \cite{Angulo1999,Le2003} that aim to improve the prediction of \emph{all} types of coexisting phenomena, for which existing active learning algorithms for sampling measurements only from the target phenomenon  can be extended and applied straightforwardly, as detailed in Section \ref{method}.}.

To achieve this, we model all types of coexisting phenomena (i.e., target and auxiliary) jointly as a \emph{multi-output Gaussian process}
%\footnote{One may argue for a simpler alternative of using the measurements of the auxiliary phenomena as additional input features to a single-output GP modeling the target phenomenon. This is, however, not feasible in practice: Such measurements have to be known/specified for GP prediction, which is not the case since they need to be sampled, just like that of the target phenomenon.} 
(MOGP) 
%Vasudevan2011,Osborne2008,,Williams2009
\cite{Alvarez2011,Bonilla2008,Teh2005}, which allows the spatial correlation structure of each type of phenomenon and the cross-correlation structure between different types of phenomena to be formally characterized.
More importantly, unlike the non-probabilistic multivariate regression methods,  
%(e.g., multivariate linear regression \cite{Reinsel1998}, multi-output support vector regression, regularization methods \cite{Evgeniou2004}),
%Izenman1975, \cite{Sanchez2004}
the probabilistic MOGP regression model allows the predictive uncertainty of the target phenomenon (as well as the auxiliary phenomena) to be formally quantified (e.g., based on entropy or mutual information criterion) and consequently exploited for deriving the active learning criterion.

To the best of our knowledge, this paper is the first to present an efficient algorithm for active learning of a MOGP model. We consider utilizing the entropy criterion to measure the predictive uncertainty of a target phenomenon, which is widely used for active learning of a single-output GP model.
Unfortunately, for the MOGP model, such a criterion scales poorly in the number of candidate sampling locations of the target phenomenon (Section~\ref{method}) and even more so in the number of selected observations (i.e., sampling budget) when optimized (Section~\ref{m.greedy}). To resolve this scalability issue, we first exploit a structure common to a unifying framework of sparse MOGP models (Section~\ref{CGP}) for deriving a novel active learning criterion (Section~\ref{method}). Then, we define a relaxed notion of submodularity\footnote{The original notion of submodularity has been used in \cite{Krause2012,Guestrin08} to theoretically guarantee the performance of their algorithms for active learning of a \emph{single-output} GP model.} called $\epsilon$-submodularity and exploit the $\epsilon$-submodularity property of our new criterion for devising a polynomial-time approximation algorithm that guarantees a constant-factor approximation of that achieved by the optimal set of selected observations (Section \ref{m.greedy}).
%Krause2007,
We empirically evaluate the performance of our proposed algorithm using three real-world datasets (Section~\ref{experiment}).
%
% whose time complexity is independent of the domain size during execution	
%This paper presents an optimal algorithm for non-myopic MOAL based on the predictive uncertainty of MOGP. 
%Unfortunately, the entropy criterion which is commonly used in active learning/sensing problem is very expensive to compute in our asymmetric multi-output system because it depends on the size of the entire environment (Section \ref{m.entropy}). Therefore, we exploit the sparse structure of MOGP model to construct a novel and efficient active learning criterion whose computational complexity is independent of the environmental size (Section \ref{m.sparse}). 
%More importantly, for the optimization problem with the new criterion, we can find a tractable greedy approximation algorithm which scales better than the optimal one in the total number of observations (Section \ref{m.greedy}). Furthermore, the concept of \textit{submodularity} which has been used to develop performance bound for different problems is relaxed to \textit{$\epsilon$-submodularity}, such that the performance of our greedy algorithm can be lower bounded (Section \ref{m.bound}). We empirically evaluate the performance of our proposed approximation algorithm using two real-world datasets (Section~\ref{experiment}).
%
\section{Modeling Coexisting Phenomena with Multi-Output Gaussian Process (MOGP)} \label{CGP}\label{PITC}
%Wackernagel98,Cressie1993,
\subsubsection{Convolved MOGP (CMOGP) Regression.} A number of MOGP models such as co-kriging \cite{Webster01}, parameter sharing \cite{Skolidis2012}, and \emph{linear model of coregionalization} (LMC) \cite{Teh2005,Bonilla2008} have been proposed to handle multiple types of correlated outputs. A generalization of LMC called the \emph{convolved MOGP} (CMOGP) model has been empirically demonstrated in \cite{Alvarez2011} to outperform the others and will be the MOGP model of our choice due to its approximation whose structure can be exploited for deriving our active learning criterion and in turn an efficient approximation algorithm, as detailed later.
%  which is used in our problem was shown .
%, especially if one output type is a blurred version of the others.
%Besides, it is easy to construct the correlation between different output types by exploiting convolution, which is the main issue for multi-output GP regression.  

Let $M$ types of coexisting phenomena be defined to vary as a realization of a CMOGP over a domain corresponding to a set $D$ of sampling locations such that each location $x \in D$ is associated with noisy realized (random) output measurement $y_{\langle x, i\rangle}$ ($Y_{\langle x, i\rangle}$) if $x$ is observed (unobserved) for type $i$ for $i=1,\ldots,M$. 
Let $D_i^+ \triangleq \{\langle x, i\rangle\}_{ x \in D}$ and $D^+ \triangleq \bigcup_{i=1}^M D_i^+$. Then, measurement $Y_{\langle x, i\rangle}$ of type $i$ is defined as a convolution between a
%In an environment with $m$ types of measurements, let $D$ representing the domain of the locations, 
%a tuple $\langle x, i\rangle$ with $ x \in D$ and $i=1, ..., m$ representing a location for measurement type $i$, $D_i' \triangleq \{\langle x, i\rangle\}_{ x \in D}$, and $D' \triangleq \cup_{i=1}^m D_i'$ such that each $\langle x, i\rangle \in D'$ can be associated with a realized (random) measurement/output $y_{\langle x, i\rangle}$ ($Y_{\langle x, i\rangle}$).
%Each type of measurement is considered as one type of output in the multi-output regression model.
%To capture the correlation between different output types, 
%consider $m$ functions for $m$ types of correlated outputs,
%In a CGP model, the function for output type $i$ is expressed as a convolution between a 
smoothing kernel $K_i( x)$ and a latent measurement function $L( x)$\footnote{To ease exposition, we consider a single latent function. Note, however, that multiple latent functions can be used to improve the fidelity of modeling, as shown in \cite{Alvarez2011}. More importantly, our proposed algorithm and theoretical results remain valid with multiple latent functions.} corrupted by an additive noise $\varepsilon_i\sim \mathcal{N}(0, \sigma^2_{n_i})$ with noise variance $\sigma^2_{n_i}$:
$$Y_{\langle x, i\rangle} \triangleq \int_{x'\in D}K_i( x-x')\ L(x')\ \text{d}x'+\varepsilon_i\ .
$$
%It has been shown that if the latent function $f_0( x)$ is a GP, the function $f_i( x)$ is also a GP.
%Let $Y_{\langle x, i\rangle} = f_i( x)+w_i$, where $w_i \sim \mathcal{N}(0, \sigma^2_{n_i})$ and $\sigma^2_{n_i}$ is the noise variance, 
As shown in \cite{Alvarez2011}, if $\{L( x)\}_{x\in D}$ is a GP, then $\{Y_{\langle x,i\rangle}\}_{\langle x,i\rangle \in D^+}$ is also a GP, that is, every finite subset of $\{Y_{\langle x,i\rangle}\}_{\langle x,i\rangle \in D^+}$ follows a multivariate Gaussian distribution. Such a GP is fully specified by its \emph{prior} mean $\mu_{\langle x,i\rangle} \triangleq \mathbb{E}[Y_{\langle x,i\rangle}]$ and covariance $\sigma_{\langle x,i\rangle\langle x',j\rangle} \triangleq$ cov$[Y_{\langle x, i\rangle}, Y_{\langle x', j\rangle}]$ for all $\langle x, i\rangle, \langle x', j\rangle\in D^+$, the latter of which characterizes the spatial correlation structure for each type of phenomenon (i.e., $i=j$) and the cross-correlation structure between different types of phenomena (i.e., $i \neq j$). 
Specifically, let $\{L( x)\}_{x\in D}$ be a GP with prior covariance $\sigma_{xx'} \triangleq \mathcal{N}( x - x'| \underline{0}, P_0^{-1})$ and $K_i( x) \triangleq \sigma_{s_i}\mathcal{N}( x|\underline{0}, P^{-1}_i)$ where $\sigma^2_{s_i}$ is the signal variance controlling the intensity of measurements of type $i$, $P_0$ and $P_i$ are diagonal precision matrices controlling, respectively, the degrees of correlation between latent measurements
 and cross-correlation between latent and type $i$ measurements, and $\underline{0}$ denotes a column vector comprising components of value $0$. Then, 
\begin{equation}\label{kernel}
\sigma_{\langle x,i\rangle\langle x',j\rangle} \hspace{-1mm}=\hspace{-0.5mm} \sigma_{s_i}\sigma_{s_j}\mathcal{N}( x -  x'| \underline{0}, P_0^{-1}+P^{-1}_i+P^{-1}_j)+\delta^{ij}_{ x  x'}\sigma^2_{n_i} 
\end{equation}
where $\delta^{ij}_{ x  x'}$ is a Kronecker delta of value $1$ if $i=j$ and $x= x'$, and 0 otherwise.

%As these $m$ types of measurements may be partially observed for each location $ x$ in a heterotopic system, 
Supposing a column vector $y_X$ of realized measurements is available for some set $X \triangleq \bigcup_{i=1}^M X_i$ of tuples of observed locations and their corresponding measurement types where $X_i \subset D_i^+$,
%Let $X_i \subseteq D_i^+$ denote a set of tuples of observed locations and their corresponding measurement type $i$, $X \triangleq \bigcup_{i=1}^M X_i$, and $y_X \triangleq (y_{\langle x, i\rangle})^\top_{\langle x, i\rangle \in X}$ be a column vector of their corresponding realized measurements.
a CMOGP regression model can exploit these observations to provide a Gaussian predictive distribution $\mathcal{N}(\mu_{Z|X},\Sigma_{ZZ|X})$ of the measurements for any set $Z \subseteq D^+ \setminus X$ of tuples of unobserved locations and their corresponding measurement types
%to predict the measurement for any set of unobserved locations $Z \subseteq D' \setminus X$. The joint distribution of $Y_Z$ will be Gaussian 
with the following \emph{posterior} mean vector and covariance matrix:
\begin{equation} \label{var}
\begin{array}{rl}
\mu_{Z|X}\hspace{-2.8mm} &\triangleq  \mu_Z + \Sigma_{ZX}\Sigma_{XX}^{-1}({y}_{X}-\mu_X)\vspace{0.5mm}\\
\Sigma_{ZZ|X} \hspace{-2.8mm}&\triangleq  \Sigma_{ZZ} - \Sigma_{ZX}\Sigma_{XX}^{-1}\Sigma_{XZ}
\end{array}
\end{equation}
where $\Sigma_{AA'} \triangleq (\sigma_{\langle x, i\rangle\langle x', j\rangle})_{\langle x, i\rangle \in A,\langle x', j\rangle \in A'}$ and $\mu_A \triangleq (\mu_{\langle x, i\rangle})^\top_{\langle x, i\rangle \in A}$ for any $A,A'\subseteq D^+$.
% and $\sigma_{\langle x, i\rangle\langle x', j\rangle }$ is computed with $k_{ij}( x,  x')$ in \eqref{kernel}.
%Note that the posterior variance \eqref{var} is independent of the observed measurements $\mathbf{y}_{X}$ , which is an important property of GP and will be used to do the active sensing work in Section \ref{method}.
%
%$\Sigma_{ZX}$ is the covariance matrix evaluated at every pair of locations in $Z$ and $X$ with each entry to be computed with \eqref{kernel}, $\Sigma_{XZ}$ is the transpose of $\Sigma_{ZX}$, and $\Sigma_{XX}$ $(\Sigma_{ZZ})$ is the covariance matrix evaluated at every pair of locations in $X$ ($Z$). Note that the posterior variance \eqref{var} is independent of the observed measurements $Y_{X}$ , which is an important property of GP and will be used to do the active sensing work in Section \ref{method}.
%
%\noindent
\subsubsection{Sparse CMOGP Regression.}
A limitation of the CMOGP model is its poor scalability in the number $|X|$ of observations:
Computing its Gaussian predictive distribution \eqref{var} requires inverting $\Sigma_{XX}$, which incurs $\mathcal{O}(|X|^3)$ time.
% For $m$ types of outputs, each having $n$ observations, the prediction requires inverting the covariance matrix $\Sigma_{XX}$, which incurs $O(m^3n^3)$ time and $O(m^2n^2)$ memory. 
To improve its scalability, a unifying framework of sparse CMOGP regression models such as the deterministic training conditional, fully independent training conditional, and \emph{partially independent training conditional} (PITC) approximations \cite{Alvarez2011} exploit a vector $L_U \triangleq (L(x))^\top_{x \in U}$ of inducing measurements for some small set $U \subset D$ of inducing locations
%\footnote{The procedure to select $U$ will be described in Section~\ref{experiment}.}
%several approximation methods have been proposed \cite{Alvarez2011}.
%supposing the latent function $f_0( x)$ is evaluated at a small set of locations $U \subset D$ and we use $\mathbf{f}_U \triangleq (f_0( x))^\top_{ x \in U}$ to represent a vector of latent variables evaluated at $U$.
(i.e., $|U| \ll |D|$) to approximate each measurement $Y_{\langle x,i\rangle}$:
$$
Y_{\langle x,i\rangle} \approx \int_{x'\in D}K_i( x-x')\ \mathbb{E}[L(x')|L_U]\ \text{d}x' + \varepsilon_i\ .
$$
They also share two structural properties that can be exploited for deriving our active learning criterion and in turn an efficient approximation algorithm: ({\bf P1}) Measurements of different types (i.e., $Y_{D^+_i}$ and $Y_{D^+_j}$ for $i\neq j$) are conditionally independent given $L_U$, and ({\bf P2}) $Y_X$
and $Y_Z$ are conditionally independent given $L_U$. 
PITC will be used as the sparse CMOGP regression model in our work here since the others in the unifying framework impose further assumptions.
%A family of approximation methods for CGP follows from these assumptions . 
With the above structural properties, PITC can utilize the observations to provide a Gaussian predictive distribution $\mathcal{N}(\mu^\text{\tiny{PITC}}_{Z|X},\Sigma^\text{\tiny{PITC}}_{ZZ|X})$ where
%As the first two methods add additional assumptions, we will use PITC as the underlying model in this work. 
%Then, the  of the joint predictive distribution become:
\begin{equation}\label{PITC.var}
\begin{array}{rl}
\mu^\text{\tiny{PITC}}_{Z|X}\hspace{-2.8mm} &\triangleq \mu_Z + \Gamma_{ZX}(\Gamma_{XX}+\Lambda_X)^{-1}(y_{X}-\mu_X)\vspace{0.5mm}\\
\Sigma^\text{\tiny{PITC}}_{ZZ|X}\hspace{-2.8mm}  &\triangleq \Gamma_{ZZ} + \Lambda_{Z} - \Gamma_{ZX}(\Gamma_{XX}+\Lambda_{X})^{-1}\Gamma_{XZ}
\end{array}
\end{equation}
such that $\Gamma_{AA'} \triangleq \Sigma_{AU}\Sigma^{-1}_{UU}\Sigma_{UA'}$  for any $A,A'\subseteq D^+$, $\Sigma_{AU}$ ($\Sigma_{UU}$) is a covariance matrix with covariance components
$\sigma_{\langle x,i\rangle x'}=\sigma_{s_i}\mathcal{N}( x -  x'| \underline{0}, P_0^{-1}+P^{-1}_i)$ for all $\langle x,i\rangle\in A$ and $x'\in U$ ($\sigma_{xx'}$ for all $x,x'\in U$), 
$\Sigma_{UA'}$ is the transpose of $\Sigma_{A'U}$,
and $\Lambda_A$ is a block-diagonal matrix constructed from the $M$ diagonal blocks of $\Sigma_{AA|U} \triangleq \Sigma_{AA}-\Gamma_{AA}$ for any $A\subseteq D^+$, each of which is a matrix $\Sigma_{A_iA_i|U}$ for $i=1,\ldots,M$ where $A_i\subseteq D^+_i$ and $A \triangleq \bigcup_{i=1}^M A_i$.
Note that computing \eqref{PITC.var} does not require the inducing locations $U$ to be observed. Also, the covariance matrix $\Sigma_{XX}$ in \eqref{var} is approximated by a reduced-rank matrix $\Gamma_{XX}$ summed with the resulting sparsified residual matrix $\Lambda_X$. So, by using the matrix inversion lemma to invert the approximated covariance matrix $\Gamma_{XX} + \Lambda_X$ and applying some algebraic manipulations,
computing \eqref{PITC.var} incurs $\mathcal{O}(|X|(|U|^2+(|X|/M)^2))$ time   \cite{Alvarez2011} in the case of $|U|\leq |X|$ and evenly distributed observations among  all $M$  types.
%
%depending on the choice of $U$, it incurs $O(mn^3)$ or $O(|U|^2mn)$ time and $O(mn^2+mn|U|)$ memory which is linear in $m$  
%Notice that, except for reducing the computational complexity of the prediction, these sparse assumptions are also important for giving efficient MOAL criteria (Section \ref{method}) and deriving tractable approximation algorithm (Section \ref{m.greedy}).
%
\section{Active Learning of CMOGP} \label{method}
%Although the sparse approximation method has reduced the computational complexity of the regression model
%In many real world applications, the number of observations we can use for prediction is limited due to limited resources (e.g., budget, time, electrical energy, etc). With a budget of $k$ sensors, it requires us to select the most informative measurements to observe.
%
%\subsection{MOAL with entropy} \label{m.entropy}
Recall from Section~\ref{sect:intro} that the entropy criterion can be used to measure the predictive uncertainty of the unobserved areas of a target phenomenon.
Using the CMOGP model (Section~\ref{CGP}), the Gaussian posterior joint entropy (i.e., predictive uncertainty) of the measurements $Y_Z$ for any set $Z \subseteq D^+ \setminus X$ of tuples of unobserved locations and their corresponding measurement types can be expressed in terms of its posterior covariance matrix $\Sigma_{ZZ|X}$ \eqref{var} which is independent of the realized measurements $y_X$:
%Entropy is commonly used to measure the uncertainty of the prediction. With the CGP model in Section \ref{CGP}, any set of unobserved outputs $Y_Z$ can be formulated in one single GP and the posterior covariance matrix $\Sigma_{ZZ|X}$ in \eqref{var}, which is independent of the measurement $\mathbf{y}_X$, can be used to quantify the uncertainty of the prediction by computing the joint entropy
$$
H(Y_Z|Y_X) \triangleq \frac{1}{2} \log (2\pi e)^{|Z|}|\Sigma_{ZZ|X}|\ .
$$
%In our problem, the system is \textit{asymmetric} because only the predictive performance of the primary output type is important. 
Let index $t$ denote the type of measurements of the target phenomenon\footnote{Our proposed algorithm can be extended to handle multiple types of target phenomena, as  demonstrated in Section~\ref{experiment}.}.
Then, active learning of a CMOGP model involves selecting an optimal set $X^*\triangleq  \bigcup_{i=1}^M X_i^*$ of $N$ tuples (i.e., sampling budget) of sampling locations and their corresponding measurement types to be observed that minimize the posterior joint entropy of type $t$ measurements at the remaining unobserved locations of the target phenomenon: 
\begin{equation}\label{h}
X^*  \triangleq \mathop{\arg\min}_{X:|X|=N} H(Y_{V_t \setminus X_t} | Y_X)
\end{equation}
where $V_t \subset D_t^+$ is a finite set of tuples of candidate sampling locations of the target phenomenon and their corresponding measurement type $t$ available to be selected for observation.
%to get observations for output type $t$.
However, evaluating the $H(Y_{V_t \setminus X_t} | Y_X)$ term in \eqref{h} incurs $\mathcal{O}(|V_t|^3 + N^3)$ time, which is prohibitively expensive when the target phenomenon is spanned by a large number $|V_t|$ of candidate sampling locations. 
If auxiliary types of phenomena are missing or ignored (i.e., $M=1$), then such a computational difficulty can be eased instead by solving the well-known \emph{maximum entropy sampling} (MES) problem \cite{Shewry87}: $X^*_t = \arg\max_{X_t:|X_t|=N} H(Y_{X_t})$ which can be proven to be equivalent to \eqref{h} by using the chain rule for entropy $H(Y_{V_t})=H(Y_{X_t}) + H(Y_{V_t \setminus X_t} | Y_{X_t})$ and noting that $H(Y_{V_t})$ is a constant.
Evaluating the $H(Y_{X_t})$ term in MES incurs $\mathcal{O}(|X_t|^3)$ time, which is independent of $|V_t|$. Such an equivalence result can in fact be extended and applied to minimizing the predictive uncertainty of \emph{all} $M$ types of coexisting phenomena, as  exploited by multivariate spatial sampling algorithms \cite{Angulo1999,Le2003}:
%Le1994,
\begin{equation}\label{h1}
\mathop{\arg\max}_{X:|X|=N} H(Y_X) = \mathop{\arg\min}_{X:|X|=N} H(Y_{V \setminus X} | Y_X),
\end{equation}
where $V \triangleq \bigcup_{i=1}^M V_i$ and $V_i$ is defined in a similar manner to $V_t$ but for measurement type $i\neq t$. This equivalence result \eqref{h1} also follows from the chain rule for entropy  $H(Y_{V})=H(Y_{X}) + H(Y_{V \setminus X} | Y_{X})$ and the fact that $H(Y_{V})$ is a constant.
Unfortunately, it is not straightforward to derive such an equivalence result for our active learning problem \eqref{h} in which a target phenomenon of interest coexists with auxiliary types of phenomena (i.e., $M > 1$):
%
%An important issue for solving \eqref{h} is that the computational complexity of $H(V_t \setminus X_t | X)$ highly depends on $|V_t|$ as it requires computing the determinant of a matrix with size $|V_t \setminus X_t| \times |V_t \setminus X_t|$ which incurs $O(|V_t \setminus X_t|^3)$ time. If the candidate locations $V_t$ are densely selected and the sensors are sparsely deployed as in most real world applications, it will be expensive to compute the entropy criterion.
%Existing symmetric MOAL problem \cite{Le1994,Le2003} which aimed to reduce the predictive uncertainty of all output types has a similar computational issue for their criterion $H(Y_{V \setminus X} | Y_X)$ and used the chain rule of entropy to solve it:
%
% Selecting the observations with maximal joint entropy will naturally minimize the predictive uncertainty of the unobserved measurements because the sum of two terms in \eqref{h1} is $H(Y_V)$ which is a constant for Gaussian entropy. Furthermore, the computational complexity of $H(Y_X)$ is independent of $|V|$.
%However, in our problem, it is not straightforward to formulate a maximization problem equivalent to \eqref{h} whose criterion computation is independent of $|V_t|$ using the same trick as \eqref{h1}. 
If we consider maximizing $H(Y_X)$ or $H(Y_{X_t})$, then it is no longer equivalent to minimizing $H(Y_{V_t \setminus X_t}|Y_X)$ \eqref{h} as the sum of the two entropy terms is not necessarily a constant.

\subsubsection{Exploiting Sparse CMOGP Model Structure.} 
%\label{m.sparse}
We derive a new equivalence result by considering instead a constant entropy $H(Y_{V_t}|L_U)$ that is conditioned on the inducing measurements $L_U$ used in sparse CMOGP regression models (Section~\ref{PITC}). 
Then, by using the chain rule for entropy and structural property {\bf P2} shared by sparse CMOGP regression models in the unifying framework \cite{Alvarez2011} described in Section~\ref{PITC}, \eqref{h} can be proven\if\myproof1 (see Appendix~\ref{a0}) \fi\if\myproof0 \cite{AA16} \fi to be equivalent to 
%Then, by adding additional terms and using the chain rule of entropy, the minimization problem in \eqref{h} can be reformulated as
%
%Unfortunately, computing expression \eqref{h2} still incurs $O(|V_t|^3)$ time. To make the computation to be independent of $|V_t|$, we exploit the sparse CGP assumptions which are used in DTC, FITC and PITC as shown in Section \ref{PITC}. With assumption (b), $I(Y_{V_t \setminus X_t}; Y_X| \mathbf{f}_U)$ goes to 0, and $H(Y_{X_t}|\mathbf{f}_U, Y_{V_t \setminus X_t})$ becomes $H(Y_{X_t}|\mathbf{f}_U)$.
%Then, problem in \eqref{h} can be approximated by solving the following problem:
\begin{equation}\label{h3}
X^* \triangleq \mathop{\arg\max}_{X:|X|=N} H(Y_{X_t}|L_U) - I(L_U;Y_{V_t \setminus X_t}|Y_X)
\end{equation}
where 
\begin{equation}\label{h3.1}
\hspace{-0.1mm}I(L_U;Y_{V_t \setminus X_t}|Y_X) \triangleq H(L_U|Y_X) - H(L_U|Y_{X\cup {V_t \setminus X_t}})\hspace{-1.6mm}
\end{equation}
is the conditional mutual information between $L_U$ and $Y_{V_t \setminus X_t}$ given $Y_X$. 
Our novel active learning criterion in \eqref{h3} exhibits an interesting exploration-exploitation trade-off: The inducing measurements $L_U$ can be viewed as latent structure of the sparse CMOGP model to induce conditional independence properties {\bf P1} and {\bf P2}.
%that is used to transfer information between according to structural 
So, on one hand, maximizing the $H(Y_{X_t}|L_U)$ term aims to select tuples $X_t$ of locations with the most uncertain measurements $Y_{X_t}$ of the target phenomenon and their corresponding type $t$ to be observed given the latent model structure $L_U$ (i.e., exploitation).
On the other hand, minimizing the $I(L_U;Y_{V_t \setminus X_t}|Y_X)$ term \eqref{h3.1} aims to select tuples $X$ to be observed  (i.e., possibly of measurement types $i\neq t$) so as to rely less on measurements $Y_{V_t \setminus X_t}$ of type $t$ at the remaining unobserved locations of the target phenomenon to infer  latent model structure $L_U$ (i.e., exploration) since $Y_{V_t \setminus X_t}$ won't be sampled.
%
% between maximizing the $H(Y_{X_t}|L_U)$ term and minimizing the $I(L_U;Y_{V_t \setminus X_t}|Y_X)$ term. 
 %As the latent function can be treated as the model structure in a CGP model, it is used to transfer information between the observed and unobserved part according to sparse assumption (b). 
% Therefore, the first term aims to select the most uncertain primary measurements to observe given the current model $\mathbf{f}_U$ (i.e., exploitation), and the second term aims to select $X$ such that we rely less on the unobserved measurements to infer the latent structure $\mathbf{f}_U$ (i.e., exploration).

Supposing $|U|\leq |V_t|$, evaluating our new active learning criterion in \eqref{h3} incurs $\mathcal{O}(|U|^3+N^3)$ time for every $X\subset V$ and a \emph{one-off} cost of 
$\mathcal{O}(|V_t|^3)$ time\if\myproof1 (Appendix~\ref{a1}). \fi\if\myproof0 \cite{AA16}. \fi
In contrast, computing the original criterion in \eqref{h} requires $\mathcal{O}(|V_t|^3 +N^3)$ time for every $X\subset V$, which is more costly, especially when the number $N$ of selected observations is much less than the number $|V_t|$ of candidate sampling locations of the target phenomenon due to, for example, a tight sampling budget or a large sampling domain that usually occurs in practice.
%
%In this new criterion, the first term can be computed in $O(|U|^3+|X_t||U|^2+|X_t|^3)$ time. The second term incurs $O(|U|^3+|X||U|^2+|X|^3)$ time during execution with assumption (a) although a one-time off preprocessing cost of $O(|V_t|^3)$ is needed prior (Appendix \ref{a1}). 
%Note that similar trick can also be used to reduce the computational complexity of $\eqref{greedy}$ in the next section.
%As the observations and the latent function are usually sparsely sampled, the computational complexity of \eqref{h3} is expected to be much better than that of $\eqref{h}$. 
%
The trick to achieving such a computational advantage can be inherited by our approximation algorithm to be described next.
 %
%exploring the exact structure of latent function and exploiting the most uncertainty locations given $\mathbf{f}_U$. The exploitation part is straightforward to understand by maximizing $H(Y_{X_t}|\mathbf{f}_U)$. For the exploration of $\mathbf{f}_U$, we need to select the observations which can best evaluate the latent process $\mathbf{f}_U$ so that the unobserved part will not give much more information about it (i.e., minimize $I(\mathbf{f}_U;Y_{V_t \setminus X_t}|Y_X)$). More importantly, with the sparse assumptions, we can find a tractable greedy algorithm for the problem in \eqref{h3} as will be shown in the next section.
%
%More importantly, we find an approximation algorithm for \eqref{h2} whose performance is bounded with a constant factor.  
%
%two issues for active sensing which are naturally addressed by equation \eqref{h2}:
%(a) Given $\mathbf{f}_U$, we need to sample the most uncertain locations for the primary output type  (i.e., maximize $H(X_t|\mathbf{f}_U)$), and
%(b) we need to select the observations which can best evaluate the latent process $\mathbf{f}_U$ so that the unobserved part will not give much more information about it (i.e., minimize $I(\mathbf{f}_U;Y_{V_t \setminus X_t}|X)$).
%As the observations and the latent process are usually sparsely sampled, the computational complexity of \eqref{h2} should be much lower than \eqref{h1}. Details about the complexity analysis will be shown in Section \ref{cc}. 
%More importantly, we find an approximation algorithm for \eqref{h2} whose performance is bounded with a constant factor.  
%
\section{Approximation Algorithm} \label{m.greedy}
Our novel active learning criterion in \eqref{h3}, when optimized, still suffers from poor scalability in the number $N$ of selected observations (i.e., sampling budget) like the old criterion in \eqref{h} because it involves evaluating a prohibitively large number of candidate selections of sampling locations and their corresponding measurement types (i.e., exponential in $N$).
%still suffers from the exponential number of possible selections problem, just like the old criterion. 
However, unlike the old criterion, it is possible to devise an efficient approximation algorithm with a theoretical performance guarantee to optimize our new criterion, which is the main contribution of our work in this paper.

%In the previous section, an optimal MOAL solution for asymmetric system has been proposed. However, if the total number of observations to be selected is large, it is still expensive to solve the problem in \eqref{h3} as the number of possible joint observations grows exponentially in $k$. 
%To solve this issue, we propose a tractable greedy approximation algorithm whose computation is linear in $k$ by exploiting sparse CGP assumptions. 
%More importantly, it has a bounded performance compared with the optimal solution in last section.

The key idea of our proposed approximation algorithm is to greedily select the next tuple of sampling location and its corresponding measurement type to be observed that maximally increases our criterion in \eqref{h3}, and iterate this till $N$ tuples are selected for observation.
Specifically, 
%The greedy algorithm will add the observations in sequence and choose the next best observation which gives maximal increase in the criterion \eqref{h3}. 
let
\begin{equation}\label{yehong}
\hspace{-0.1mm}
F(X) \triangleq H(Y_{X_t}|L_U)-I(L_U;Y_{V_t \setminus X_t}|Y_X) + I(L_U; Y_{V_t})
\hspace{-2mm}
\end{equation}
denote our active learning criterion in \eqref{h3} augmented by a positive constant 
%$C = H(\mathbf{f}_U)-H(\mathbf{f}_U|Y_{V_t})$ 
$I(L_U; Y_{V_t})$
to make $F(X)$ non-negative. 
Such an additive constant $I(L_U; Y_{V_t})$ is simply a technical necessity for proving the performance guarantee and does not affect the outcome of the optimal selection (i.e., $X^* = \arg\max_{X:|X|=N} F(X)$).
%Note that the constant term is required to give a performance guarantee and will not change the results in both optimal and approximate algorithms. 
Then, our approximation algorithm greedily selects the next tuple $\langle x,i\rangle$ of sampling location $x$ and its corresponding measurement type $i$ that maximizes $F(X \cup\{ \langle x, i\rangle\}) - F(X)$:
\begin{equation}\label{greedy}
\hspace{-1.8mm}
\begin{array}{rl}
\langle x, i\rangle^+ 
\hspace{-3.1mm}&\triangleq\hspace{-1mm} \displaystyle \mathop{\arg\max}_{\langle x, i\rangle \in V \setminus X} F(X \cup\{ \langle x, i\rangle\}) - F(X) \\
& \displaystyle=\hspace{-1mm}\mathop{\arg\max}_{\langle x, i\rangle \in V \setminus X} H(Y_{\langle x, i\rangle}|Y_X)\hspace{-0.5mm}-\delta_i H(Y_{\langle x, i\rangle}|Y_{X\cup V_t \setminus X_t})
\end{array}
\end{equation}
where $\delta_i$ is a Kronecker delta of value $0$ if $i = t$, and $1$ otherwise. The derivation of~\eqref{greedy} is in\if\myproof1 Appendix~\ref{a2}. \fi\if\myproof0 \cite{AA16}. \fi Our algorithm updates $X \leftarrow X \cup \{\langle x, i\rangle^+\}$ and iterates the greedy selection \eqref{greedy} and the update till $|X|=N$ (i.e., sampling budget is depleted).
The intuition to understanding \eqref{greedy} is that our algorithm has to choose between observing a sampling location with the most uncertain measurement (i.e., $H(Y_{\langle x, t\rangle}|Y_X)$) of the target phenomenon (i.e., type $t$) vs. that for an auxiliary type $i\neq t$ inducing the largest reduction in predictive uncertainty of the
%sharing the most amount of information $I(Y_{\langle x, i\rangle }; Y_{V_t \setminus X_t}|Y_X)$ with 
measurements at the remaining unobserved locations of the target phenomenon since $H(Y_{\langle x, i\rangle}|Y_X)- H(Y_{\langle x, i\rangle}|Y_{X\cup V_t \setminus X_t})=I(Y_{\langle x, i\rangle }; Y_{V_t \setminus X_t}|Y_X) = H(Y_{V_t \setminus X_t}|Y_X)- H(Y_{V_t \setminus X_t}|Y_{X\cup \{\langle x, i\rangle\}})$.

It is also interesting to figure out whether our approximation algorithm may avoid selecting tuples of a certain auxiliary type $i\neq t$ and formally analyze the conditions under which it will do so, as elucidated in the following result:
\begin{proposition} \label{lemma.g}
Let $V_{\text{-}t}\triangleq\bigcup_{i\neq t} V_i$, $X_{\text{-}t}\triangleq\bigcup_{i\neq t} X_i$, $\rho_i \triangleq \sigma^2_{s_i} / \sigma^2_{n_i}$, and $R(\langle x, i\rangle,V_t \setminus X_t) \triangleq\sum_{\langle x', t\rangle \in V_t \setminus X_t}\mathcal{N}( x -  x'| \underline{0}, P_0^{-1}+P^{-1}_i+P^{-1}_t)^2$. Assuming absence of suppressor variables, $H(Y_{\langle x, i\rangle }|Y_X)-H(Y_{\langle x, i\rangle }|Y_{X\cup V_t \setminus X_t}) \leq 0.5\log (1+4\rho_t\rho_iR(\langle x, i\rangle,V_t \setminus X_t))$ for any $\langle x,i\rangle\in V_{\text{-}t} \setminus X_{\text{-}t}$.
%$\langle x,i\rangle\in V\setminus  (X\cup V_t \setminus X_t)$.
\end{proposition}
Its proof\if\myproof1 (Appendix~\ref{a2.1}) \fi\if\myproof0 \cite{AA16} \fi 
%For the primary output type $t$, it will select the most uncertain measurement to observe; For the auxiliary output types, it will select the auxiliary observations which shares the most information with the unobserved part of the primary output given the current observations as the criterion becomes a mutual information $I(Y_{\langle x, i\rangle }; Y_{V_t \setminus X_t}|Y_X)$.
relies on the
following assumption of the absence of suppressor variables which holds in many practical cases  \cite{Das2008}: Conditioning does not make $Y_{\langle x,i\rangle}$ and $Y_{\langle x',t\rangle}$ more correlated for any $\langle x,i\rangle\in  V_{\text{-}t} \setminus X_{\text{-}t}$ and $\langle x',t\rangle\in V_t\setminus X_t$.
	%which has been shown to be true in many real world problems \cite{Das2008}.
Proposition~\ref{lemma.g} reveals that when the signal-to-noise ratio $\rho_i$ of auxiliary type $i$ is low (e.g., poor-quality measurements due to high noise) and/or the cross correlation \eqref{kernel} between measurements of the target phenomenon and auxiliary type $i$ is small due to low $\sigma^2_{s_t}\sigma^2_{s_i}R(\langle x, i\rangle,V_t \setminus X_t) $, our greedy criterion in \eqref{greedy} returns a small value, hence causing
our algorithm to avoid selecting tuples of auxiliary type $i$.
\begin{theorem}[Time Complexity]\label{thm:time}
Our approximation algorithm incurs $\mathcal{O}(N(|V||U|^2+N^3)+|V_t|^3)$ time.
\end{theorem}
Its proof is in\if\myproof1 Appendix~\ref{a2.5}. \fi\if\myproof0 \cite{AA16}. \fi 
So, our approximation algorithm only incurs quartic time in the number $N$ of selected observations and cubic time in the number $|V_t|$ of candidate sampling locations of the target phenomenon.
\subsubsection{Performance Guarantee.} %\label{m.bound}
To theoretically guarantee the performance of our approximation algorithm, we will first motivate the need to define a relaxed notion of \emph{submodularity}.
%Next, we will use the \emph{submodularity} property of a set function to give a performance guarantee for . 
A submodular set function exhibits a natural diminishing returns property: When adding an element to its input set, the increment in its function value decreases with a larger input set.
%In the context of work here, adding data to a set of observations gives less information than adding it to its subset.
%Formally,
%$$
%g(A' \cup \{a\}) - g(A') \leq g(A \cup \{a\}) - g(A)
%$$
%for every $A \subseteq A'$.
To maximize a nondecreasing and submodular set function, the work of \citeauthor{Nemhauser1978}~\shortcite{Nemhauser1978} has proposed a greedy algorithm  guaranteeing a $(1-1/e)$-factor approximation of that achieved by the optimal input set.
%proven that the greedy algorithm is guaranteed to have a performance of $(1-1/e)OPT$, where $OPT$ is the value of $g(A^*)$ for the optimal result $A^*$.

The main difficulty in proving the submodularity of $F(X)$ \eqref{yehong} lies in its mutual information term being conditioned on $X$.
%submodularity of criterion $F(X)$ which involves a mutual information term conditioned on $X$ is not obvious. 
Some works \cite{Krause05,Renner2002} have shown the submodularity of such conditional mutual information by imposing conditional independence assumptions (e.g., Markov chain). 
In practice, these strong assumptions (e.g., $Y_A \perp Y_{\langle x, i\rangle} |Y_{V_t \setminus X_t}$ for any $A \subseteq X$ and $\langle x,i\rangle\in V_{\text{-}t} \setminus X_{\text{-}t}$) severely violate the correlation structure of multiple types of coexisting phenomena  and are an overkill: The correlation structure can in fact be preserved to a fair extent 
%too strong as we want to preserve the correlations between all observations to do prediction. 
by relaxing these assumptions, which consequently entails a relaxed form of submodularity of $F(X)$ \eqref{yehong}; a performance guarantee similar to that of \citeauthor{Nemhauser1978}~\shortcite{Nemhauser1978} can then be derived for our approximation algorithm.
\begin{definition}
	A function G : $2^B \rightarrow \mathbb{R}$ is $\epsilon$-submodular if 
	$
	G(A' \cup \{a\}) - G(A') \leq G(A \cup \{a\}) - G(A) + \epsilon
	$
for any $A \subseteq A' \subseteq B$ and $a \in B \setminus A'$.
\end{definition}
%Then, let $A \subseteq X$ be a subset of $X$ and $\sigma^2_{n_{min}} \triangleq \min_{i=1}^m \sigma^2_{n_i}$, we have the following property for our criterion $F(X)$:
\begin{lemma}\label{near-s}
Let $\sigma^2_{n^{*}} \triangleq \min_{i\in\{1,\ldots,M\}}\sigma^2_{n_i}$.
Given $\epsilon_1\geq 0$, if
%If there exists some $\epsilon_1\geq 0$ such that 
\begin{equation}\label{sert}
\Sigma^\text{\tiny{\emph{PITC}}}_{\langle x, i\rangle\langle x, i\rangle  | \widetilde{X}\cup V_t \setminus X_t} - \Sigma^\text{\tiny{\emph{PITC}}}_{\langle x, i\rangle\langle x, i\rangle  | X\cup V_t \setminus X_t} \leq \epsilon_1
\end{equation}
for any $\widetilde{X} \subseteq X$ and $\langle x, i\rangle  \in V_{\text{-}t} \setminus X_{\text{-}t}$, then $F(X)$ is $\epsilon$-submodular where $\epsilon = 0.5 \log(1+ \epsilon_1/\sigma^2_{n^*})$. 
\end{lemma}
Its proof is in\if\myproof1 Appendix~\ref{a3}. \fi\if\myproof0 \cite{AA16}. \fi 
Note that \eqref{sert} relaxes the above example of conditional independence assumption (i.e., assuming $\epsilon_1=0$) to one which allows $\epsilon_1>0$.
In practice, $\epsilon_1$ is expected to be small:
% and should have small  in the real problems:
%The variance reduction term $\sigma^2_{\langle x, i\rangle |V_t \setminus X_t} - \sigma^2_{\langle x, i\rangle |V_t \setminus X_t, A}$ will be small because the size of $V_t \setminus X_t$ is usually very large.
Since further conditioning monotonically decreases a posterior variance \cite{Xu2014}, an expected large set $V_t \setminus X_t$ of tuples of remaining unobserved locations of the target phenomenon tends to be informative enough to make 
$\Sigma^\text{\tiny{{PITC}}}_{\langle x, i\rangle\langle x, i\rangle  | \widetilde{X}\cup V_t \setminus X_t}$ small and hence the non-negative variance reduction term and $\epsilon_1$ in \eqref{sert} small.

%It has been shown that the posterior variance of GP is monotonically decreasing and lower bounded by $\sigma^2_{n_{min}}$ \cite{Cao2013}. Intuitively,
% if the set $V_t \setminus X_t$ is large and gives enough information to make $\sigma^2_{\langle x, i\rangle |V_t \setminus X_t}$ to be small, the additional set $A$ will not add much more information and the variance reduction will be small as well. 

Furthermore, 
%for any small $\epsilon_1 > 0$, 
\eqref{sert} with a given small $\epsilon_1$ can be realized by controlling the discretization of the domain of candidate sampling locations.
For example, by refining the discretization of $V_t$ (i.e., increasing $|V_t|$),
the variance reduction term in \eqref{sert} decreases because it has been shown in  \cite{Das2008} to be submodular in many practical cases.
We give another example in Lemma~\ref{lemma} to realize \eqref{sert} by controlling the discretization such that every pair of selected observations are sufficiently far apart.
% the discretization of the location space is well controlled. For example, if we keep increasing the discretization level of $V_t$, this variance reduction term will decease because it has been shown to be submodular in many practical cases \cite{Das2008}. 
%One method to realize this assumption by controlling the discretization width of the samples will be shown later in 

It is easy to derive that $F(\emptyset) = 0$. 
%H(L_U) - H(L_U|Y_{V_t}) + H(L_U) - H(L_U|Y_{V_t}).
The ``information never hurts'' bound for entropy \cite{Cover91} entails a nondecreasing $F(X)$:
$F(X \cup \{\langle x, i\rangle\}) - F(X)= H(Y_{\langle x, i\rangle }|Y_X)-\delta_i H(Y_{\langle x, i\rangle }|Y_{X\cup V_t \setminus X_t}) \geq H(Y_{\langle x, i\rangle }|Y_X)-H(Y_{\langle x, i\rangle }|Y_{X\cup V_t \setminus X_t}) \geq 0$.
The first inequality requires $\sigma^2_{n^*} \geq (2 \pi e)^{-1}$ so that $H(Y_{\langle x, i\rangle }|Y_A) = 0.5 \log 2\pi e\Sigma^\text{\tiny{{PITC}}}_{\langle x, i\rangle\langle x, i\rangle |A} \geq  0.5 \log 2\pi e\sigma^2_{n^*} \geq 0$,\footnote{$\Sigma^\text{\tiny{{PITC}}}_{\langle x, i\rangle\langle x, i\rangle |A}\geq \sigma^2_{n^*}$ is proven in\if\myproof1  Lemma~\ref{lemma1} in Appendix~\ref{a2.1}. \fi\if\myproof0 \cite{AA16}.\fi} which is reasonable in practice due to ubiquitous noise.
%the noise variance to be not very small which makes the posterior to be positive. This is reasonable in many real world problems due to the ubiquitous noise. 
%Formally, if $\sigma^2_{n^*} \geq \frac{1}{2 \pi e}$, the posterior entropy of any data point . 
Combining this result with Lemma~\ref{near-s} yields the performance guarantee:
\begin{theorem} \label{bound}
Given $\epsilon_1\geq 0$, if \eqref{sert} holds, then our approximation algorithm is guaranteed to select $X$ s.t. $F(X) \geq (1-1/e)(F(X^*)-N\epsilon)$
	where $\epsilon = 0.5 \log(1+ \epsilon_1/\sigma^2_{n^*})$.  
\end{theorem}
Its proof\if\myproof1 (Appendix~\ref{a5}) \fi\if\myproof0 \cite{AA16} \fi 
is similar to that of the well-known result of \citeauthor{Nemhauser1978}~\shortcite{Nemhauser1978} except for exploiting $\epsilon$-submodularity of $F(X)$ in Lemma~\ref{near-s} instead of submodularity.
%inequalities following from submodularity being replaced by that due to $\epsilon$-submodularity (i.e., Lemma~\ref{near-s}).
%
%The above result can be obtained by replacing the submodular inequality with the $\epsilon$-submodular one in the well-known result of \citeauthor{Nemhauser1978}~\shortcite{Nemhauser1978}. For completeness, its proof is given in .
%
%the result of \citeauthor{Nemhauser1978}~\shortcite{Nemhauser1978} and Theorem \ref{near-s}.

Finally, we present a discretization scheme that satisfies \eqref{sert}:
Let $\omega$ be the smallest discretization width of $V_i$ for $i=1,\ldots,M$. Construct a new set $V^-\subset V$ of tuples of candidate sampling locations and their corresponding measurement types such that every pair of tuples are at least a distance of $p\omega$ apart for some $p>0$; each candidate location thus has only one corresponding type.
%will show one discretization method that guarantee the variance reduction assumption in Lemma~\ref{near-s} to be true for the situation that different output types are not sampled for one location. 
Such a construction $V^-$ constrains our algorithm to select observations sparsely across the spatial domain
so that any $\langle x, i\rangle  \in V_{\text{-}t} \setminus X_{\text{-}t}$
has sufficiently many neighboring tuples of remaining unobserved locations of the target phenomenon from $V_t \setminus X_t$ to keep 
$\Sigma^\text{\tiny{{PITC}}}_{\langle x, i\rangle\langle x, i\rangle  | \widetilde{X}\cup V_t \setminus X_t}$ small and hence the variance reduction term and $\epsilon_1$ in \eqref{sert} small. Our previous theoretical results still hold if $V^-$ is used instead of $V$.
The result below gives the minimum value of $p$ to satisfy \eqref{sert}:
%
%The general idea is to sample sparsely across the environment so that for any unobserved location $\langle x, i\rangle $, there will be enough $V_t \setminus X_t$ around it to keep the variance reduction to be small. %Let $\omega$ denotes the smallest discretization width of the environment. Let the discretization width of the samples to be $p\omega$, we can get the following lemma:
%
\begin{lemma}\label{lemma}
Let $\sigma_{s^*}^2\triangleq \max_{i\in\{1,\ldots,M\}}\sigma^2_{s_i}$, 
$\ell$ be the largest first diagonal component of $P_0^{-1}+P_i^{-1}+P_j^{-1}$ for all $i,j=1,\ldots,M$, and
$\xi \triangleq \exp(-\omega^2/(2\ell))$. 
%and $p\omega$ be the discretization width of $V^-$.
Given $\epsilon_1> 0$ and assuming absence of suppressor variables,
if %there exists some $p>0$ such that 
%	$$\
%	p^2 > \max \{ \frac{a}{\log \xi}, \frac{b}{\log \xi} \}
%	$$
%	with $a = \log\left\{\frac{1}{4\sigma_{s_{max}}^2} \left( \sqrt{\frac{4\epsilon_1\sigma^2_{n_{min}}+\epsilon_1^2k}{k}} - \epsilon_1 \right)  \right\}$ and $b = \log(\frac{\eta}{2k})$.
$$	
p^2 \hspace{-0.5mm}>\hspace{-0.5mm} \log\hspace{-0.5mm}\left\{\hspace{-0.5mm}\frac{1}{2\sigma_{s^*}^2}\hspace{-0.5mm}\min\hspace{-1mm}\left(\hspace{-1mm}\frac{\sigma_{n^*}^2}{N},\frac{1}{2}\hspace{-1mm} \left( \hspace{-1mm}\sqrt{\epsilon_1^2\hspace{-0.5mm}+\hspace{-0.5mm}\frac{4\epsilon_1\sigma^2_{n^*}}{N}}\hspace{-0.5mm} - \hspace{-0.5mm}\epsilon_1\hspace{-1mm} \right)\hspace{-1mm}\right) \hspace{-1mm} \right\} \hspace{-1mm}\Bigg/\hspace{-1.5mm} \log \xi ,
$$	
then \eqref{sert} holds. See\if\myproof1 Appendix~\ref{a4} \fi\if\myproof0 \cite{AA16} \fi for its proof.
\end{lemma}
%Its proof is given in Appendix~\ref{a4}.\vspace{-5mm}
%
%\subsection{Computational complexity} \label{cc}
%
\section{Experiments and Discussion}\label{experiment}
This section evaluates the predictive performance of our approximation algorithm (m-Greedy) empirically on three real-world datasets: (a) \emph{Jura} dataset \cite{Goovaerts97b} contains concentrations of $7$ heavy metals collected at $359$ locations in a Swiss Jura region; 
(b) \emph{Gilgai} dataset \cite{Webster77} contains electrical conductivity and chloride content generated from a line transect survey of $365$ locations of Gilgai territory in New South Wales, Australia; and
(c) \emph{indoor environmental quality} (IEQ) dataset  \cite{Guestrin04} contains temperature ($^\circ $F) and light (Lux) readings taken by $43$ temperature sensors and $41$ light sensors deployed in the Intel Berkeley Research lab. 
The sampling locations for the Jura and IEQ datasets are shown in\if\myproof1 Appendix~\ref{datasets}. \fi\if\myproof0 \cite{AA16}. \fi
%Maps for dataset (a) and (c) are shown in Figure \ref{fig:map}.

The performance of m-Greedy is compared to that of the
%the random method (m-Random) and 
(a) maximum variance/entropy (m-Var) algorithm which greedily selects the next location and its corresponding measurement type with maximum posterior variance/entropy in each iteration; and
%\footnote{m-Var is equivalent to greedy algorithm with entropy criterion}; 
%For the random method, the average performance over 10 random trials is recorded. 
(b) greedy maximum entropy (s-Var) \cite{Shewry87} and mutual information (s-MI) \cite{Guestrin08} sampling algorithms for gathering observations \emph{only} from the target phenomenon.
%as the criterion, for only the target phenomena are also implemented for comparison.

For all experiments, k-means is used to select inducing locations $U$ by clustering all possible locations available to be selected for observation into $|U|$ clusters such that each cluster center corresponds to an element of $U$. The hyper-parameters (i.e., $\sigma^2_{s_i}$, $\sigma^2_{n_i}$, $P_0$ and $P_i$ for $i=1,\ldots,M$) of MOGP and single-output GP models are learned using the data via maximum likelihood estimation \cite{Alvarez2011}. 
For each dataset, observations (i.e., $100$ for Jura and Gilgai datasets and $10$ for IEQ dataset) of type $t$ are randomly selected to form the test set $T$;
the tuples of candidate sampling locations and corresponding type $t$ therefore become  less than that of auxiliary types.
%, which naturally makes the target output type to be under-sampled (i.e., less possible observations than the other output types).
%Note that we only let the algorithms choose from the locations where observations actually exist in the dataset in order to have realized observation for prediction\footnote{The algorithm and theoretical results remain valid even if we select observations from a subset of $V$.}.
The \emph{root mean squared error} (RMSE) metric $\sqrt{|T|^{-1}\sum_{x\in T} (y_{\langle x, t\rangle}-\mu_{\langle x,t\rangle|X})^2 }$ is used to evaluate the performance of the tested algorithms. All experimental results are averaged over $50$ random test sets.
For a fair comparison, the measurements of all types are normalized before using them for training, prediction, and active learning.
\subsubsection{Jura Dataset.}
Three types of correlated lg-Cd, Ni, and lg-Zn measurements are used in this experiment; we take the log of  Cd and Zn measurements to remove their strong skewness, as proposed as a standard statistical practice in \cite{Webster01}. 
The measurement types with the smallest and largest signal-to-noise ratios (respectively, lg-Cd and Ni; see\if\myproof1 Appendix~\ref{datasets2}) \fi\if\myproof0 \cite{AA16}) \fi 
are each set as type $t$.

%
%Output types with the smallest S-N ratio (log(Cd)) and the largest S-N ratio (Ni) as shown in  are respectively used as the target output for the experiments.
%
\begin{figure}
	\centering
	%\vspace{-3mm}
	\begin{tabular}{ccc}
		\hspace{-0mm}\includegraphics[scale=0.188]{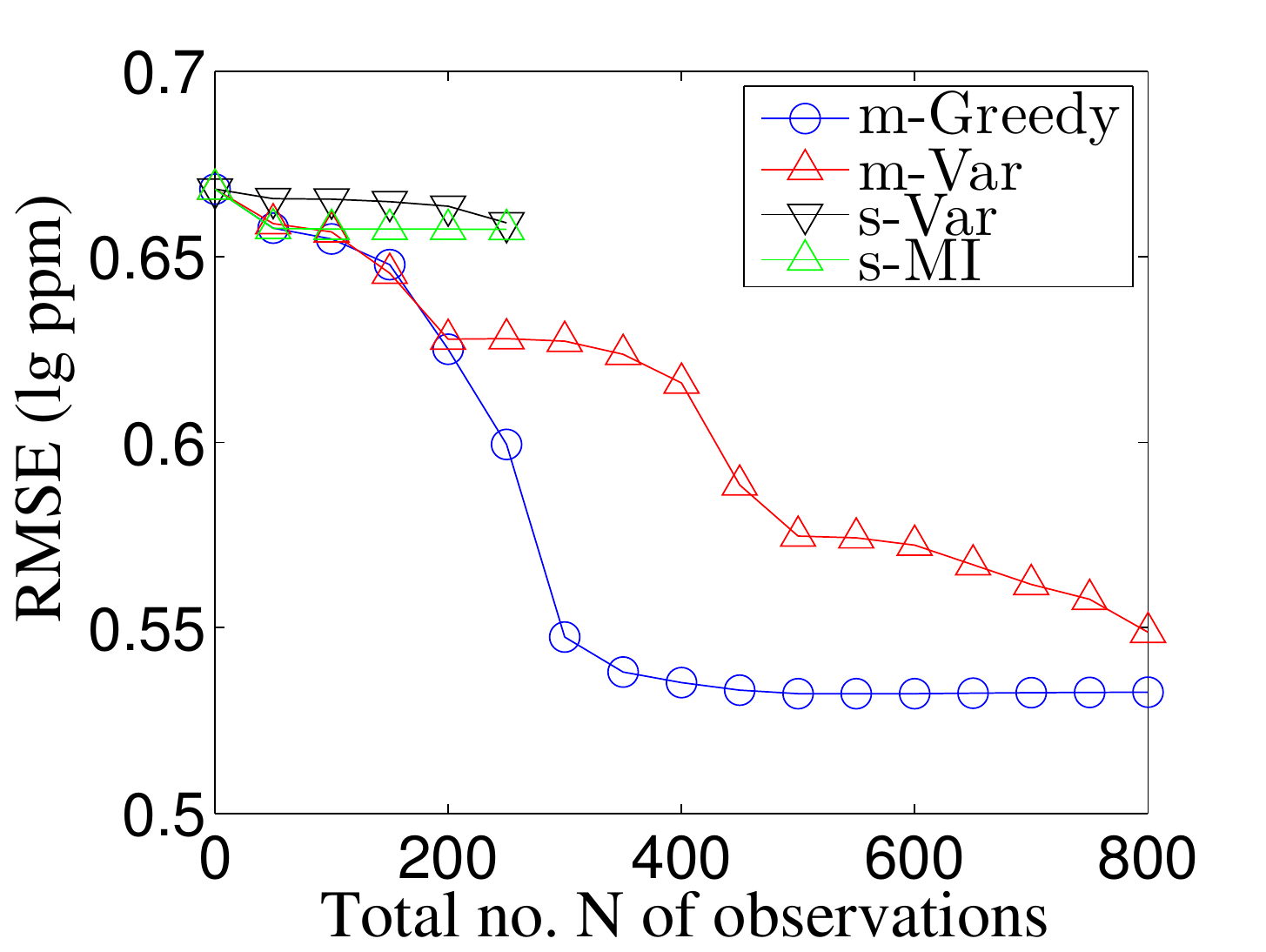} & \hspace{-4mm}\includegraphics[scale=0.188]{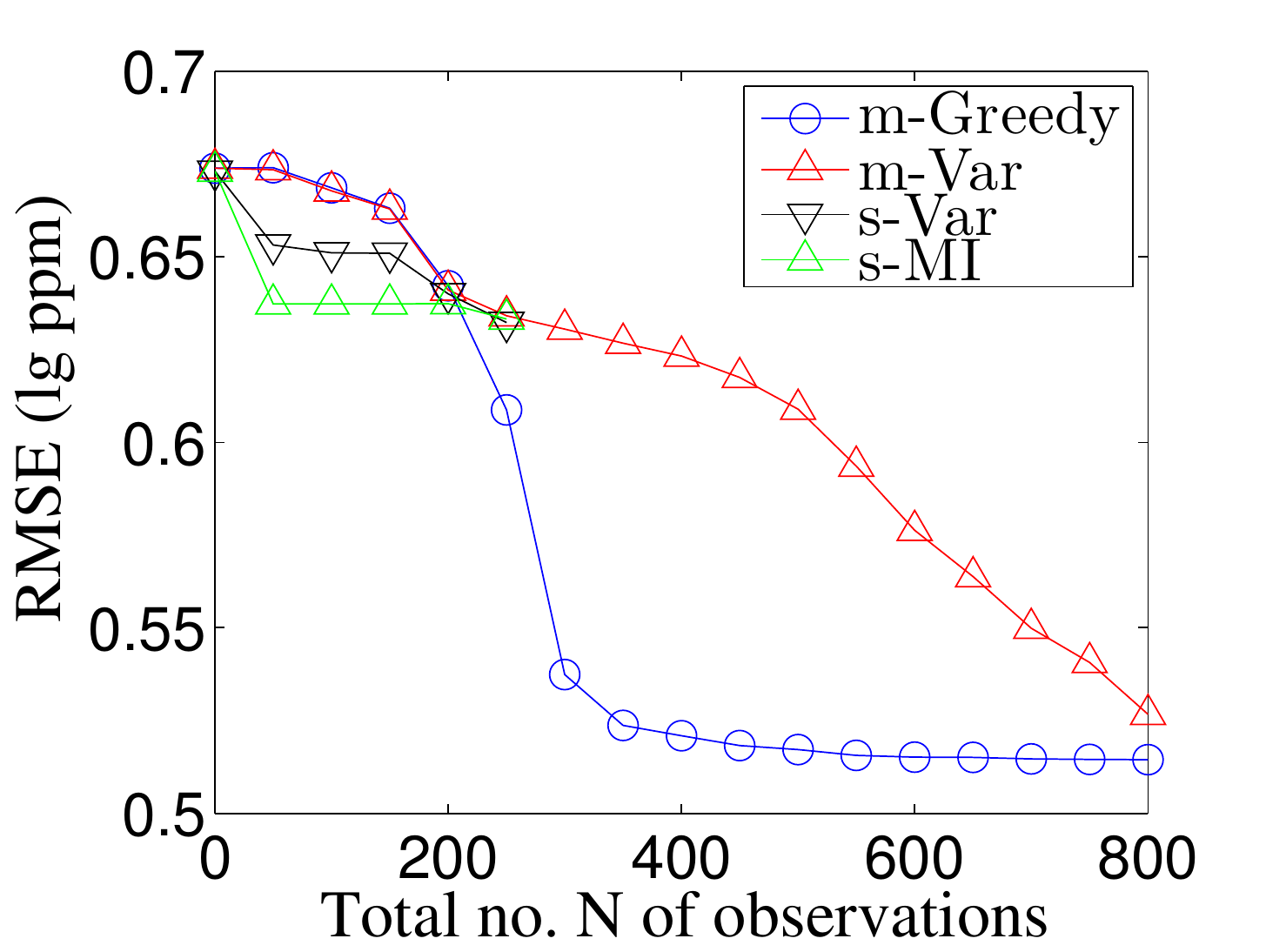} & \hspace{-4mm}\includegraphics[scale=0.188]{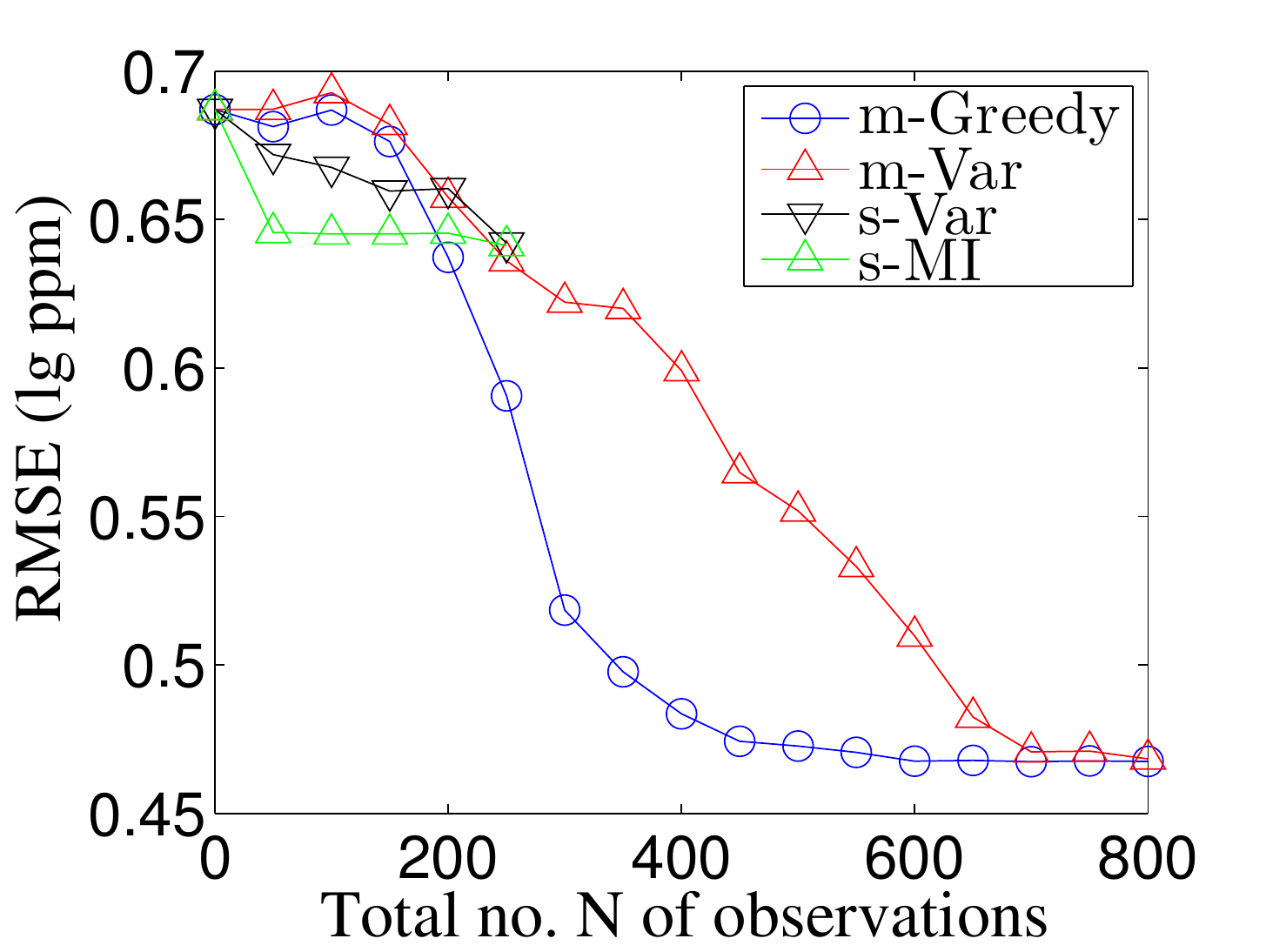} \vspace{-1.5mm}\\
		{\scriptsize (a) $|U| = 50$} & \hspace{-4mm}{\scriptsize (b) $|U| = 100$} & \hspace{-4mm}{\scriptsize (c) $|U| = 200$} \vspace{-0.5mm}\\
		\hspace{-0mm}\includegraphics[scale=0.19]{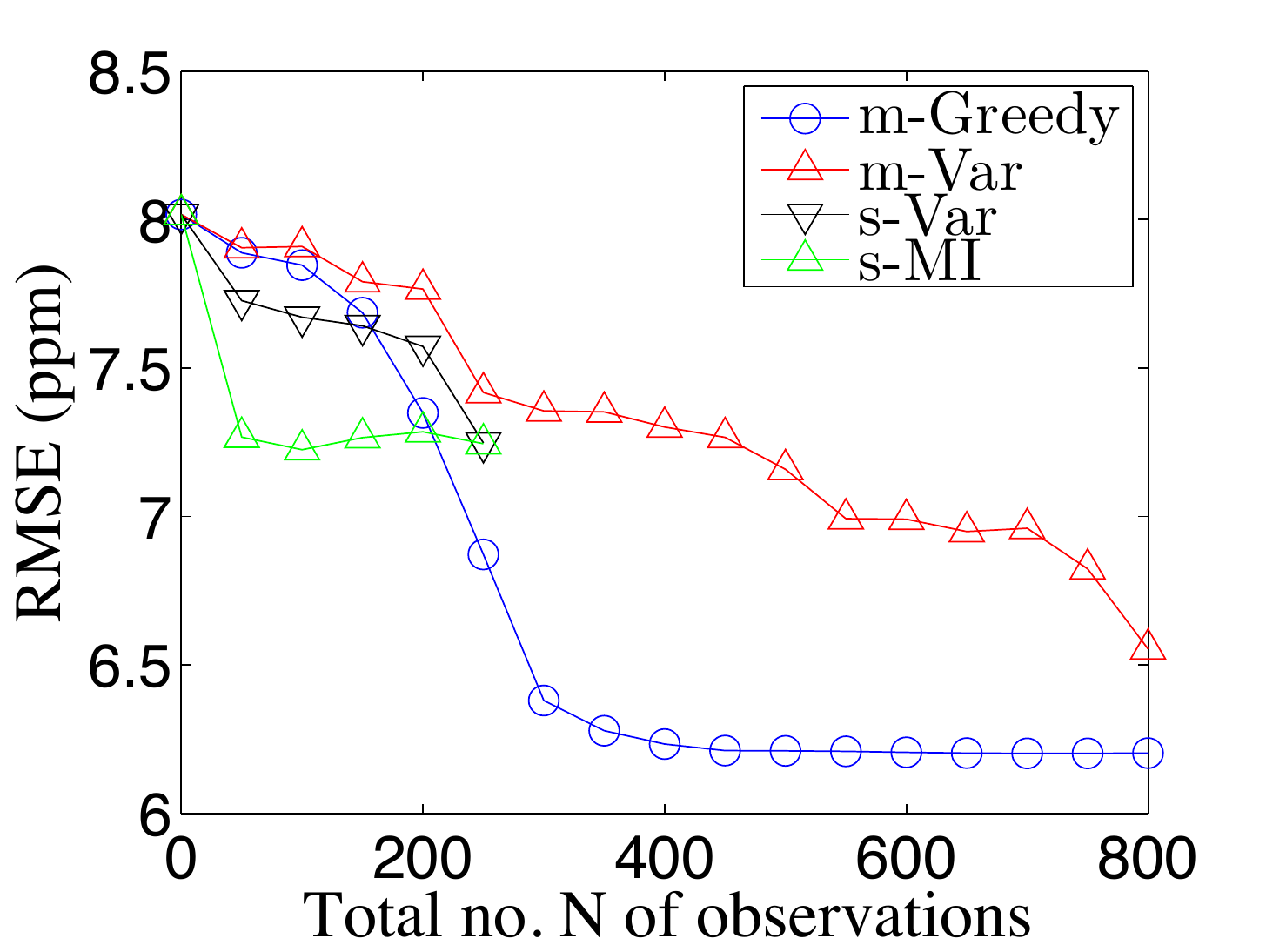} & \hspace{-4mm}\includegraphics[scale=0.19]{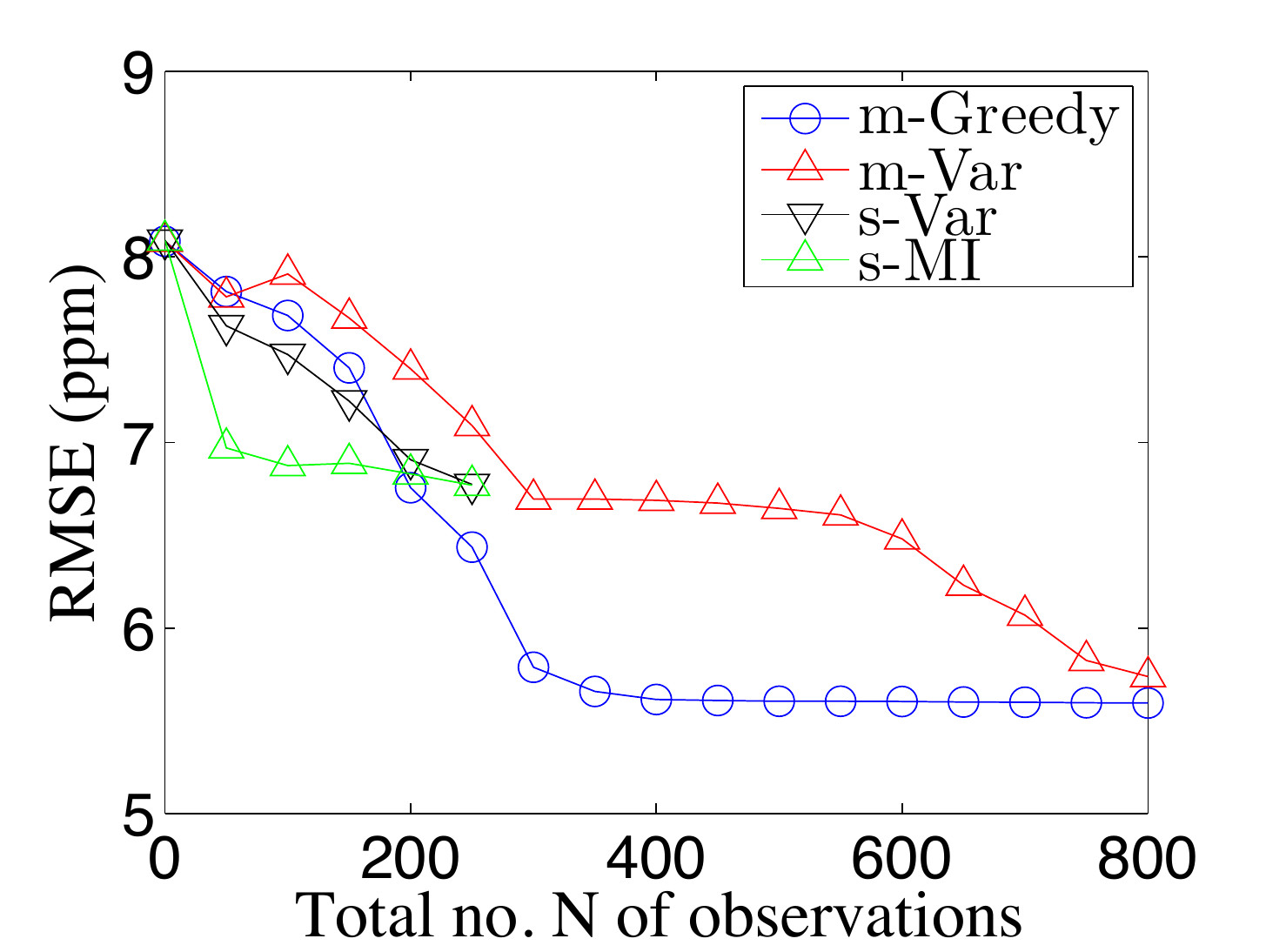} & \hspace{-4mm}\includegraphics[scale=0.19]{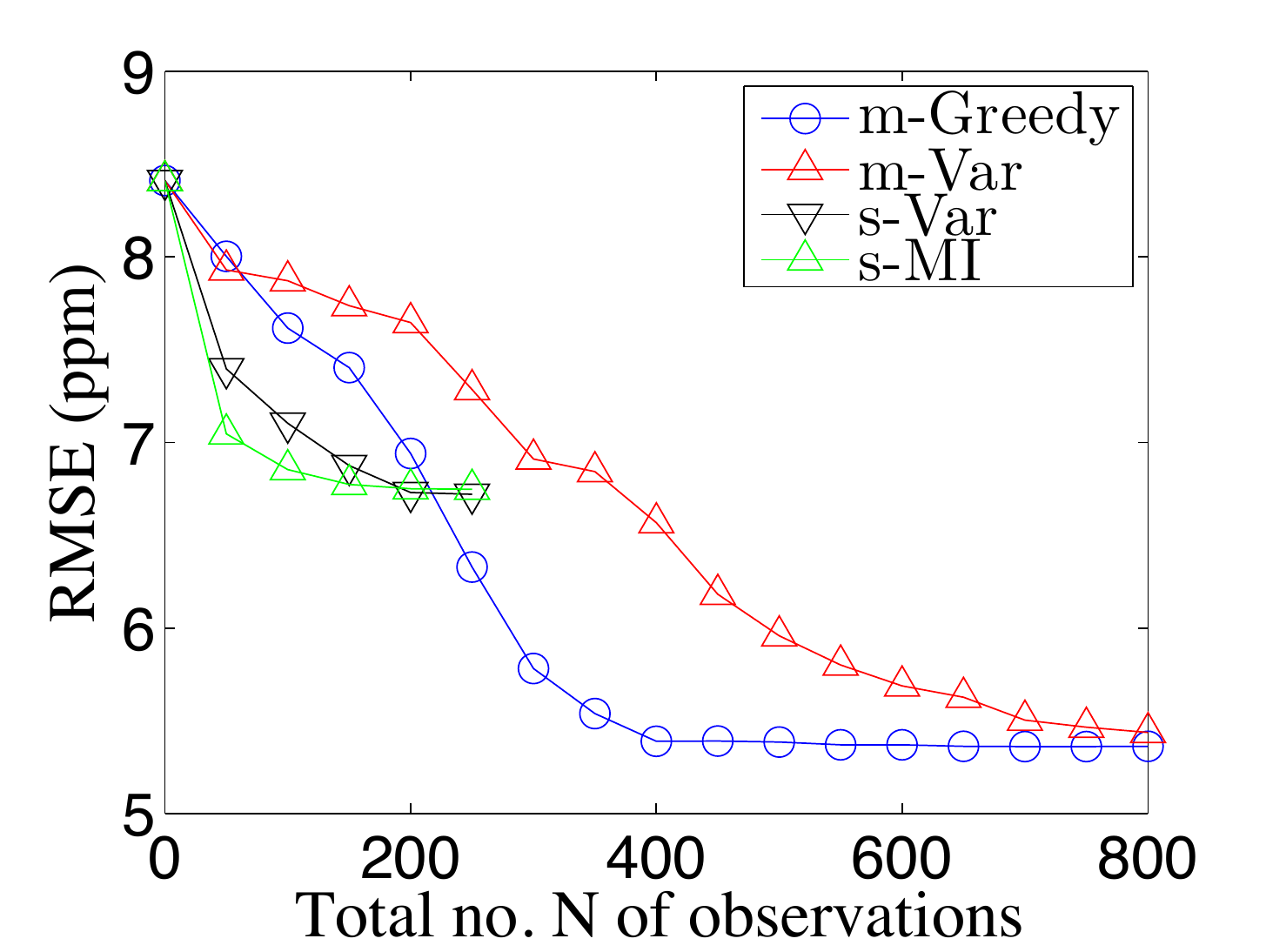} \vspace{-1.5mm}\\
		{\scriptsize (d) $|U| = 50$} & \hspace{-4mm}{\scriptsize (e) $|U| = 100$} & \hspace{-4mm}{\scriptsize (f) $|U| = 200$}\vspace{-3mm}
	\end{tabular}
	\caption{Graphs of RMSEs vs. no. $N$ of observations with (a-c) lg-Cd and (d-f) Ni as type $t$ and varying no. $|U| = 50, 100, 200$ of inducing locations for Jura dataset.\vspace{-4mm}}
	\label{fig:jura}
	%\vspace{-5mm}
\end{figure}
Figs.~\ref{fig:jura}a-c and \ref{fig:jura}d-f show, respectively, results of the tested algorithms with lg-Cd and Ni as type $t$. It can be observed that the RMSE of m-Greedy decreases more rapidly than that of m-Var, especially when observations of auxiliary types are selected after about $N=200$. 
This is because our algorithm selects observations of auxiliary types that induce the largest reduction in predictive uncertainty of the measurements at the remaining unobserved locations of the target phenomenon (Section~\ref{m.greedy}).
%This is the main advantage of our proposed algorithm which focuses on the correlations (i.e., MI) between auxiliary and target output types, which helps to minimize the predictive uncertainty of only the target output. 
In contrast, m-Var may select observations that reduce the predictive uncertainty of  auxiliary types of phenomena, which does not directly achieve the aim of our active learning problem.
With increasing $|U|$, both m-Greedy and m-Var reach smaller RMSEs, but m-Greedy 
can achieve this faster with much less observations.
%
%still converges faster than that m-Var even though the performance gaps get smaller.
%
As shown in Figs.~\ref{fig:jura}a-f, m-Greedy performs much better than s-Var and s-MI, which means observations of correlated auxiliary types can indeed be used to improve the prediction of the target phenomenon.
Finally, by comparing the results between Figs.~\ref{fig:jura}a-c and~\ref{fig:jura}d-f, the RMSE of m-Greedy with Ni as type $t$ decreases faster than that with lg-Cd as type $t$, especially in the beginning (i.e., $N\leq 200$) due to higher-quality Ni measurements (i.e., larger signal-to-noise ratio).
\subsubsection{Gilgai Dataset.}
In this experiment, the lg-Cl contents at depth $0$-$10$cm and $30$-$40$cm are used jointly as two types of target phenomena while the log of electrical conductivity, which is easier to  measure at these depths, is used as the auxiliary type. Fig. \ref{fig:sensor}a shows results of the average RMSE over the two lg-Cl types with $|U| = 100$. Similar to the results of the Jura dataset, with two types of target phenomena, the RMSE of m-Greedy still decreases more rapidly with increasing $N$ than that of m-Var and achieves a much smaller RMSE than that of s-Var and s-MI; the results of s-Var and s-MI are also averaged over two independent single-output GP predictions of lg-Cl content at the two depths.
\subsubsection{IEQ Dataset.}
Fig. \ref{fig:sensor}b shows results with light as type $t$ and $|U| = 40$. The observations are similar to that of the Jura and Gilgai datasets: RMSE of m-Greedy decreases faster than that of the other algorithms. 
More importantly, 
%detailed differences between single and multi-output methods is shown in this result. As can be seen from Fig. \ref{fig:sensor}b, 
with the same number of observations, m-Greedy achieves much smaller RMSE than s-Var and s-MI that can sample only from the target phenomenon.
This is because m-Greedy selects observations of the auxiliary type (i.e., temperature) that are less noisy  ($\sigma^2_{n_i} = 0.13$) than that of light ($\sigma^2_{n_t} = 0.23$), which demonstrates its advantage over s-Var and s-MI when type $t$ measurements are noisy (e.g., due to poor-quality sensors).\vspace{1mm}
%

%\hspace{-2.5mm}
\floatbox[{\capbeside\thisfloatsetup{capbesideposition={center,center},capbesidewidth=3.35cm}}]{figure}[\FBwidth]
{\caption{Graphs of RMSEs vs. no. $N$ of observations with (a) lg-Cl as types $t$ for Gilgai dataset and (b) light as type $t$ for IEQ dataset.}\label{fig:sensor}}
{\hspace{-12mm}
\begin{tabular}{cc}
		\hspace{-0mm}\includegraphics[scale=0.166]{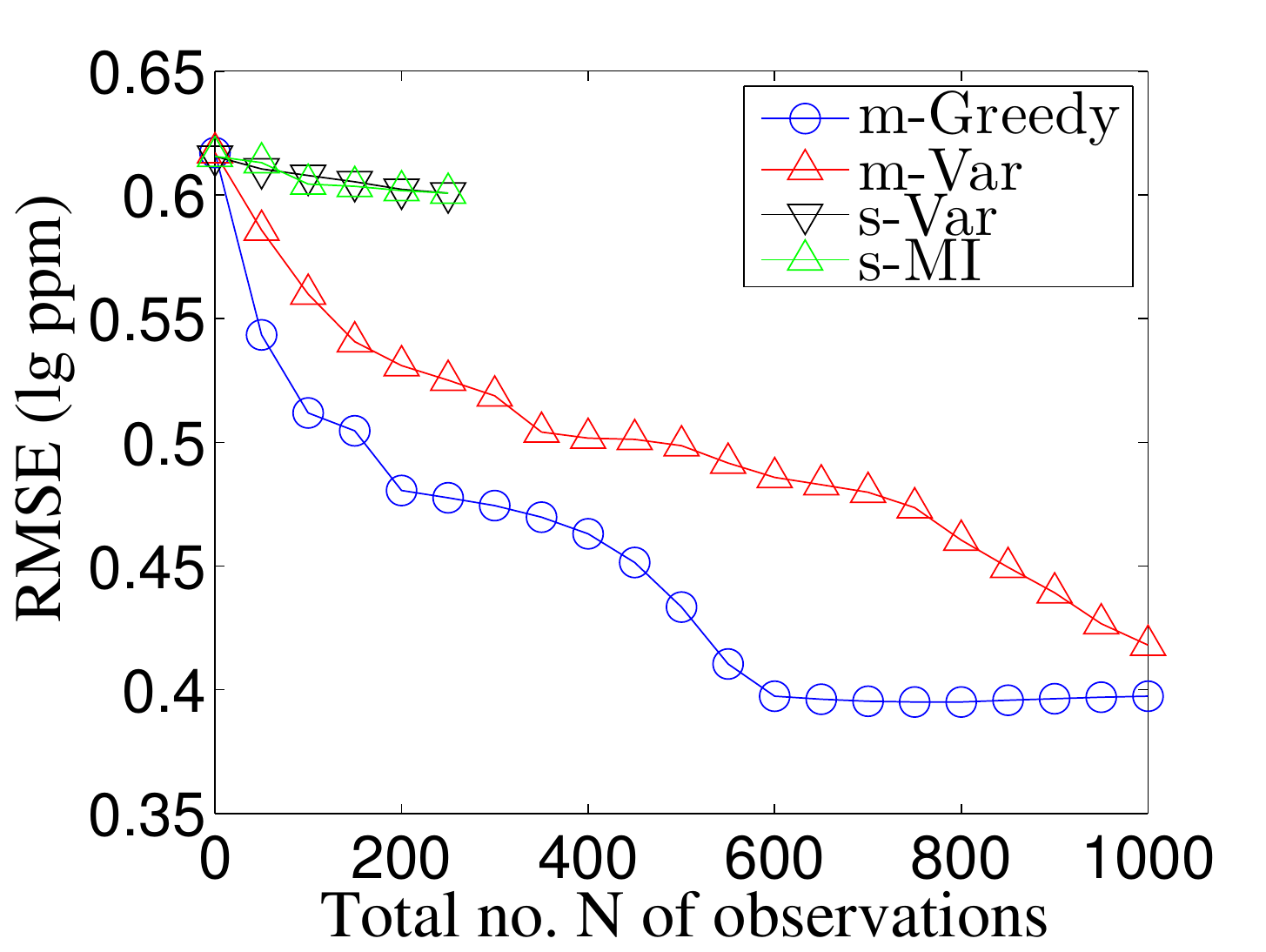} & \hspace{-4mm}\includegraphics[scale=0.166]{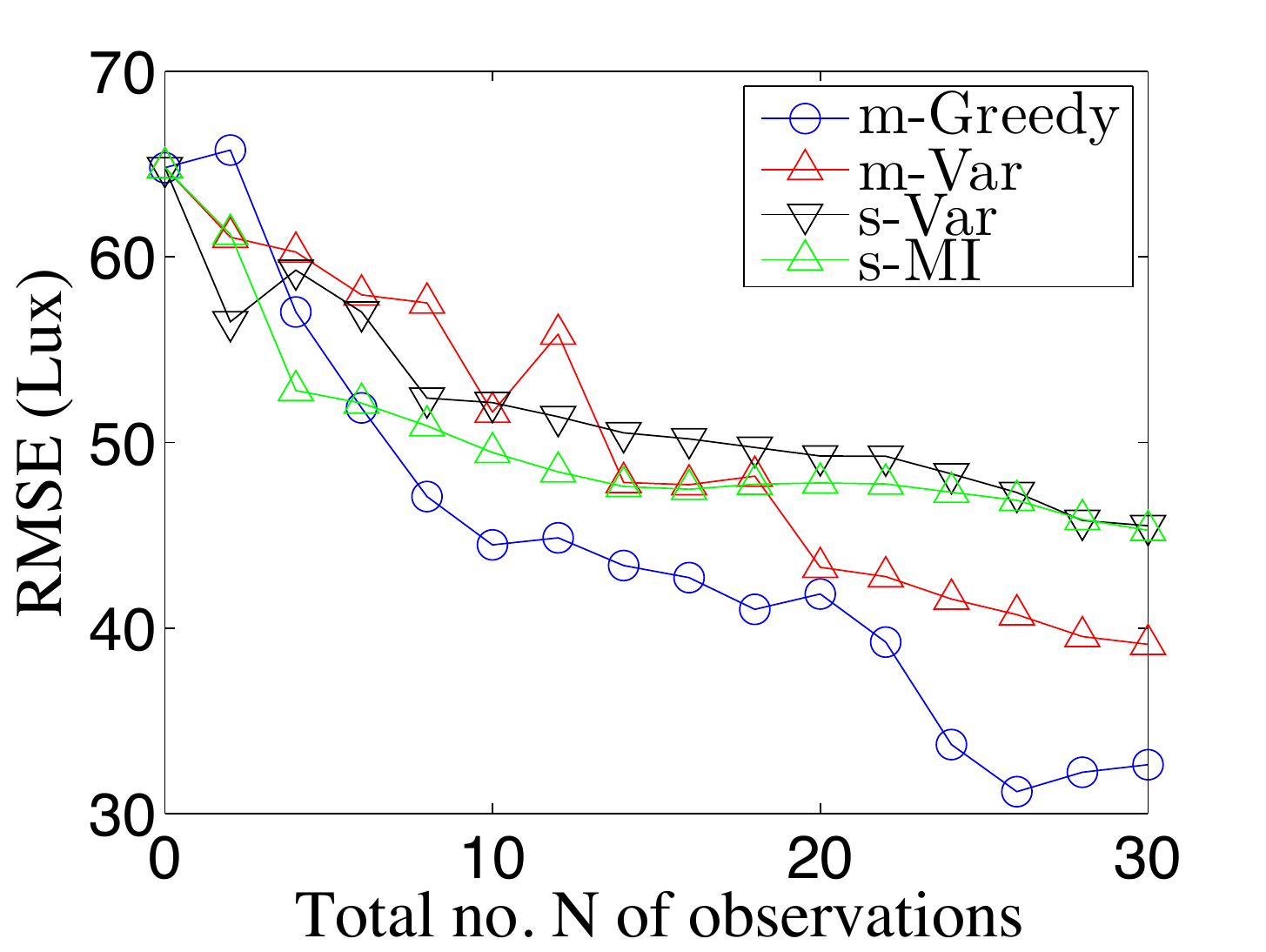} \vspace{-2mm}\\
		{\scriptsize (a)} & \hspace{-4mm}{\scriptsize (b)}
\end{tabular}\vspace{-0mm}
}\vspace{-3.5mm}

\section{Related Work}
%To the best of our knowledge, we are the first to do active learning of MOGP model. All
Existing works on active learning with multiple output measurement types are not driven by the MOGP model and have not
% existing literatures about active learning with multiple types of measurements didn't 
formally characterized the cross-correlation structure between different types of phenomena: Some spatial sampling algorithms \cite{Bueso1998,Angulo2001} have simply modeled the auxiliary phenomenon as a noisy perturbation of the target phenomenon that is assumed to be latent, which differs from our work here.
\emph{Multi-task active learning} (MTAL) and \emph{active transfer learning} (ATL) algorithms have considered the prediction of each type of phenomenon as one task and used the auxiliary tasks to help learn the target task. But, the MTAL algorithm of \citeauthor{Zhang2010}~\shortcite{Zhang2010} requires relations between different classification tasks to be manually specified, which is highly non-trivial to achieve in practice and not applicable to MOGP regression.
%as the cross-correlation between continuous measurements can be very difficult to specify. 
Some ATL and active learning algorithms \cite{Roth2006,Zhao2013} have used active learning criteria (e.g., margin-based criterion) specific to their classification or recommendation tasks that cannot be readily tailored to MOGP regression. 
%The ATL algorithm of \citeauthor{Wang2014}~\shortcite{Wang2014} has used only a single-output GP to model and predict the offset between different tasks, which may not represent a complex cross-correlation structure well (e.g., each type of coexisting phenomena is an additive combination of blurred versions of some latent ones).
%\vspace{-1.5mm}
%Shi2008,,Zhu2011
\section{Conclusion}%\vspace{-0.5mm}
This paper describes a novel efficient algorithm for active learning of a MOGP model.
To resolve the issue of poor scalability in optimizing the conventional entropy criterion, 
we exploit a structure common to a unifying framework of sparse MOGP models for deriving a novel active learning criterion \eqref{h3}. Then, we exploit the $\epsilon$-submodularity property of our new criterion (Lemma~\ref{near-s}) for devising a polynomial-time approximation algorithm \eqref{greedy} that guarantees a constant-factor approximation of that achieved by the optimal set of selected observations (Theorem~\ref{bound}).
Empirical evaluation on three real-world datasets shows that our approximation algorithm m-Greedy outperforms existing algorithms for active learning of MOGP and single-output GP models, especially when measurements of the target phenomenon are more noisy than that of the auxiliary types.
For our future work, we plan to extend our approach by generalizing non-myopic active learning \cite{Cao2013,NghiaICML14,Ling2016,LowICAPS09,LowAAMAS08,LowAAMAS11}
of single-output GPs to that of MOGPs and improving its scalability to big data through parallelization \cite{LowUAI13,LowAAAI15}, online learning \cite{Xu2014}, and stochastic variational inference \cite{NghiaICML15}.\vspace{1mm}

\noindent
{\bf Acknowledgments.}
This research was carried out at the SeSaMe Centre. It is supported by Singapore NRF under its IRC@SG Funding Initiative and administered by IDMPO.\vspace{-1mm}

\bibliographystyle{aaai}
\bibliography{moas}

\if \myproof1
\clearpage
\appendix
\section{Derivation of Novel Active Learning Criterion \eqref{h3}} \label{a0}
$$
\begin{array}{l}
\displaystyle\mathop{\arg\min}_{X:|X|=N} H(Y_{V_t \setminus X_t} | Y_X) \\
\displaystyle = \mathop{\arg\max}_{X:|X|=N} H(Y_{V_t}|L_U) - H(Y_{V_t \setminus X_t}|Y_X) \\
\displaystyle = \mathop{\arg\max}_{X:|X|=N} H(Y_{V_t}|L_U) - H(Y_{V_t \setminus X_t}|L_U) + H(Y_{V_t \setminus X_t}|L_U)\ -\\
\displaystyle\quad H(Y_{V_t \setminus X_t}|Y_X,L_U) + H(Y_{V_t \setminus X_t}|Y_X,L_U) - H(Y_{V_t \setminus X_t}|Y_X) \vspace{1mm}\\
\displaystyle = \mathop{\arg\max}_{X:|X|=N} H(Y_{X_t}|L_U, Y_{V_t \setminus X_t}) + I(Y_{V_t \setminus X_t}; Y_X|L_U)\ -\\
\displaystyle\quad I(L_U; Y_{V_t \setminus X_t}|Y_X)\vspace{1mm}\\
\displaystyle = \mathop{\arg\max}_{X:|X|=N} H(Y_{X_t}|L_U)  - I(L_U; Y_{V_t \setminus X_t}|Y_X)\ .
\end{array}
$$
The first equality follows from the fact that $H(Y_{V_t}|L_U)$ is a constant.
The third equality is due to the chain rule for entropy $H(Y_{V_t}|L_U)=H(Y_{V_t \setminus X_t}|L_U)+H(Y_{X_t}|L_U, Y_{V_t \setminus X_t})$ as well as the definition of conditional mutual information $I(Y_{V_t \setminus X_t}; Y_X|L_U)\triangleq H(Y_{V_t \setminus X_t}|L_U)- H(Y_{V_t \setminus X_t}|Y_X,L_U)$ and $I(L_U;Y_{V_t \setminus X_t}|Y_X) \triangleq H(Y_{V_t \setminus X_t}|Y_X)- H(Y_{V_t \setminus X_t}|Y_X,L_U)$.
The last equality follows from structural property {\bf P2} shared by sparse CMOGP regression models in the unifying framework \cite{Alvarez2011} described in Section~\ref{PITC}, which results in $H(Y_{X_t}|L_U, Y_{V_t \setminus X_t}) =H(Y_{X_t}|L_U)$  and $I(Y_{V_t \setminus X_t}; Y_X| L_U) = 0$.
\section{Time Complexity of Evaluating Active Learning Criterion in \eqref{h3}} \label{a1}
%The second term of \eqref{h3} can be written as
$$H(Y_{X_t}|L_U)= \frac{1}{2} \log (2\pi e)^{|X_t|}|\Sigma_{X_tX_t|U}|$$ where
$\Sigma_{X_tX_t|U}=\Sigma_{X_tX_t}-\Sigma_{X_t U}\Sigma^{-1}_{UU}\Sigma_{UX_t}$ by definition (see last paragraph of Section~\ref{PITC}). So, evaluating $H(Y_{X_t}|L_U)$ incurs $\mathcal{O}(|U|^3+N^3)$ time for every $X\subset V$; this worst-case time complexity occurs when all the tuples in $X$ are of measurement type $t$ (i.e., $X=X_t$).
$$
\begin{array}{rl}
I(L_U;Y_{V_t \setminus X_t}|Y_X)\hspace{-2.8mm} &=\displaystyle H(L_U|Y_X) - H(L_U|Y_{X\cup V_t \setminus X_t}) \\
&=\displaystyle \frac{1}{2} \log \frac{|\Sigma_{UU|X}|}{|\Sigma_{UU|X\cup V_t \setminus X_t}|} \\
&=\displaystyle \frac{1}{2} \log \frac{|\Sigma_{UU|X}|}{|\Sigma_{UU|\bigcup_{i \neq t} X_i\cup V_t}|}
\end{array}
$$
where
\begin{equation*} 
%\label{a1.1}
\Sigma_{UU|A} = \Sigma_{UU}(\Sigma_{UU}+\Sigma_{UA} \Lambda^{-1}_{A} \Sigma_{AU})^{-1}\Sigma_{UU}
\end{equation*}
for any $A\subset D^+$, as derived in \cite{Alvarez2011}.
%\sum^M_{i=1} \Sigma{UA_i} \Sigma_{A_i A_i|U}^{-1} \Sigma{A_i U}
% |X| { (|X|/M)^2 + |U|^2 } + |U|^3
Therefore, evaluating $|\Sigma_{UU|X}|$ incurs $\mathcal{O}(|U|^3+N^3)$ time for every $X\subset V$; this worst-case time complexity occurs when all the tuples in $X$ are of one measurement type.
% |X||U|^2+

Let $A \triangleq \bigcup_{i \neq t} X_i \cup V_t$. 
%The time complexity of evaluating $|\Sigma_{UU|V_t, \bigcup_{i \neq t} X_i}| = |\Sigma_{UU|B}| $ is dominated by a $|V_t|^3$ term that is incurred due to the inversion of $\Lambda_B$. However, 
Then, by the definition of $\Lambda_{A}$ (see last paragraph of Section~\ref{PITC}),
$$
\Sigma_{UA} \Lambda^{-1}_{A} \Sigma_{AU} = \sum_{i \neq t} \Sigma_{UX_i} \Sigma_{X_i X_i|U}^{-1} \Sigma_{X_iU}+\Sigma_{UV_t} \Sigma_{V_tV_t|U}^{-1} \Sigma_{V_tU}\ .
$$
Evaluating the $\sum_{i \neq t} \Sigma_{UX_i} \Sigma_{X_i X_i|U}^{-1} \Sigma_{X_iU}$ term incurs $\mathcal{O}(|U|^3+N^3)$ time for every $X\subset V$; this worst-case time complexity occurs when all the tuples in $X$ are of one measurement type.
Note that the $\Sigma_{UV_t} \Sigma_{V_tV_t|U}^{-1} \Sigma_{V_tU}$ term remains the same for every $X\subset V$ (i.e., since it is independent of $X$) and hence only needs to be computed once in $\mathcal{O}(|V_t|^3)$ time.
Therefore, evaluating $|\Sigma_{UU|\bigcup_{i \neq t} X_i\cup V_t}|= |\Sigma_{UU|A}|$ incurs $\mathcal{O}(|U|^3+N^3)$ time for every $X\subset V$ and a \emph{one-off} cost of $\mathcal{O}(|V_t|^3)$ time.
Consequently, evaluating $I(L_U;Y_{V_t \setminus X_t}|Y_X)$ incurs $\mathcal{O}(|U|^3+N^3)$ time for every $X\subset V$ and a \emph{one-off} cost of 
$\mathcal{O}(|V_t|^3)$ time.

So, evaluating our active learning criterion in \eqref{h3} incurs $\mathcal{O}(|U|^3+N^3)$ time for every $X\subset V$ and a \emph{one-off} cost of 
$\mathcal{O}(|V_t|^3)$ time.

\section{Derivation of Greedy Criterion in \eqref{greedy}} \label{a2}
%The criteria is
%$$
%F(X) = H(Y_{X_t}|\mathbf{f}_U)-H(\mathbf{f}_U|Y_X)+H(\mathbf{f}_U|Y_X, Y_{V_t \setminus X_t}) + C.
%$$
If $i = t$, then
\begin{equation}\label{B.1}
\hspace{-1.8mm}
\begin{array}{l}
\displaystyle F(X \cup \{\langle x, t\rangle\} ) - F(X) \\
\displaystyle = H(Y_{X_t\cup \{\langle x, t\rangle\} }|L_U)\ - \\
\displaystyle  \quad (H(L_U|Y_{X\cup\{\langle x, t\rangle\} }) - H(L_U|Y_{X\cup\{\langle x, t\rangle\} \cup V_t \setminus (X_t\cup\{\langle x, t\rangle\} )}))\ -\\
\displaystyle  \quad (H(Y_{X_t}|L_U) - (H(L_U|Y_X)-H(L_U|Y_{X\cup V_t \setminus X_t}))) \\
\displaystyle = H(Y_{X_t\cup \{\langle x, t\rangle\} }|L_U) - H(Y_{X_t}|L_U)\ + \\
\displaystyle  \quad (H(L_U|Y_X) - H(L_U|Y_{X\cup\{\langle x, t\rangle\} })) \\
\displaystyle = H(Y_{\langle x, t\rangle }|Y_{X_t}, L_U) + H(Y_{\langle x, t\rangle }|Y_X) - H(Y_{\langle x, t\rangle }|Y_X,L_U) \\
\displaystyle = H(Y_{\langle x, t\rangle }|L_U) + H(Y_{\langle x, t\rangle }|Y_X) - H(Y_{\langle x, t\rangle }|L_U) \\
\displaystyle = H(Y_{\langle x, t\rangle }|Y_X)\ . 
\end{array}
\end{equation}
The first equality follows from~\eqref{h3} and~\eqref{yehong}.
The second equality is due to $H(L_U|Y_{X\cup\{\langle x, t\rangle\} \cup V_t \setminus (X_t\cup\{\langle x, t\rangle\} )}) = H(L_U|Y_{X\cup V_t \setminus X_t})$.
The third equality is due to the chain rule for entropy $H(Y_{X_t\cup \{\langle x, t\rangle\} }|L_U) = H(Y_{X_t}|L_U)+ H(Y_{\langle x, t\rangle }|Y_{X_t}, L_U)$ as well as
the definition of conditional mutual information $I(L_U;Y_{\langle x, t\rangle }|Y_X) \triangleq H(L_U|Y_X) - H(L_U|Y_{X\cup\{\langle x, t\rangle\} }) = H(Y_{\langle x, t\rangle }|Y_X) - H(Y_{\langle x, t\rangle }|Y_X,L_U)$. 
The second last equality follows from structural property {\bf P2} shared by sparse CMOGP regression models in the unifying framework \cite{Alvarez2011} described in Section~\ref{PITC}.

Otherwise (i.e., $i \neq t$),
\begin{equation}\label{B.2}
%\hspace{-1.8mm}
\begin{array}{l}
F(X \cup \{\langle x, i\rangle\} ) - F(X) \\
\displaystyle = H(Y_{X_t }|L_U)\ - \\
\displaystyle  \quad (H(L_U|Y_{X\cup\{\langle x, i\rangle\} }) - H(L_U|Y_{X\cup\{\langle x, i\rangle\} \cup V_t \setminus X_t}))\ -\\
\displaystyle  \quad (H(Y_{X_t}|L_U) - (H(L_U|Y_X)-H(L_U|Y_{X\cup V_t \setminus X_t}))) \\
\displaystyle= H(Y_{X_t}|L_U) - H(Y_{X_t}|L_U)\ +  \\
\displaystyle \quad(H(L_U|Y_X) - H(L_U|Y_{X\cup\{\langle x, i\rangle\}}))\ + \\
\displaystyle \quad H(L_U|Y_{X\cup V_t \setminus X_t\cup\{\langle x, i\rangle\} }) - H(L_U|Y_{X\cup V_t \setminus X_t}) \\
\displaystyle= H(Y_{\langle x, i\rangle }|Y_X) - H(Y_{\langle x, i\rangle }|L_U, Y_X) \ +\\
\displaystyle \quad H(Y_{\langle x, i\rangle }|Y_{X\cup V_t \setminus X_t},L_U) - H(Y_{\langle x, i\rangle }|Y_{X\cup V_t \setminus X_t}) \\
\displaystyle= H(Y_{\langle x, i\rangle }|Y_X) - H(Y_{\langle x, i\rangle }|L_U)  \\
\displaystyle \quad + H(Y_{\langle x, i\rangle }|L_U) - H(Y_{\langle x, i\rangle }|Y_{X\cup V_t \setminus X_t}) \\
= H(Y_{\langle x, i\rangle }|Y_X) - H(Y_{\langle x, i\rangle }|Y_{X\cup V_t \setminus X_t})\ . \end{array}
\end{equation}
%We can get the first equality of \eqref{B.1} because \{$Y_X, Y_{V_t \setminus X_t}$\} in $H(\mathbf{f}_U|Y_X, Y_{V_t \setminus X_t})$ remains the same set when $\langle x, i\rangle  \in V_t \setminus X_t$.
The first equality follows from~\eqref{h3} and~\eqref{yehong}.
The third equality is due to the definition of conditional mutual information $I(L_U;Y_{\langle x, i\rangle }|Y_X) \triangleq H(L_U|Y_X) - H(L_U|Y_{X\cup\{\langle x, i\rangle\}}) = H(Y_{\langle x, i\rangle }|Y_X) - H(Y_{\langle x, i\rangle }|L_U, Y_X)$ and $I(L_U;Y_{\langle x, i\rangle }|Y_{X\cup V_t \setminus X_t}) \triangleq H(L_U|Y_{X\cup V_t \setminus X_t}) - H(L_U|Y_{X\cup V_t \setminus X_t\cup\{\langle x, i\rangle\} })  = H(Y_{\langle x, i\rangle }|Y_{X\cup V_t \setminus X_t} - H(Y_{\langle x, i\rangle }|Y_{X\cup V_t \setminus X_t},L_U)$.
The second last equality follows from structural properties {\bf P1} and {\bf P2} shared by sparse CMOGP regression models in the unifying framework \cite{Alvarez2011} described in Section~\ref{PITC}.
Therefore, \eqref{greedy} results.
%The second equality in both \eqref{B.1} and \eqref{B.2} are due to the property of entropy: $H(A, B) = H(A) + H(B|A) = H(B) + H(A|B)$ such that $H(A) - H(A|B) = H(B) - H(B|A)$.
%The third equality in both \eqref{B.1} and \eqref{B.2} are due to sparse assumption (b).
%Finally, from the last line of \eqref{B.1} and \eqref{B.2}, we can get the result of \eqref{greedy}.
%
\section{Proof of Proposition~\ref{lemma.g}}\label{a2.1}
Before proving Proposition~\ref{lemma.g}, the following lemmas are needed:
\begin{lemma}\label{lemma1}
	For all $X \subset V$ and $\langle x, i\rangle\in V\setminus X$, $\Sigma^\text{\tiny{\emph{PITC}}}_{\langle x, i\rangle\langle x, i\rangle|X} \geq \sigma^2_{n_{i}}$.
\end{lemma}
Its proof follows closely to that of Lemma $6$ in \cite{Cao2013}.

\begin{lemma} \label{lemma0}
Assuming absence of suppressor variables, for all $X \subset V$ and $\langle x, i\rangle, \langle x', j\rangle\in V\setminus X$, $|\Sigma^\text{\tiny{\emph{PITC}}}_{\langle x, i\rangle \langle x', j\rangle|X} | \leq 2|\sigma_{\langle x, i\rangle \langle x', j\rangle}|$.
\end{lemma}

\noindent
\emph{Proof}. If $i = j$, then
	\begin{equation} \label{l3.1}
	|\Sigma^\text{\tiny{{PITC}}}_{\langle x, i\rangle \langle x', j\rangle }| = |\sigma_{\langle x, i\rangle \langle x', j\rangle }| \leq 2|\sigma_{\langle x, i\rangle \langle x', j\rangle}|\ .
	\end{equation}
	If $i \neq j$, then
	\begin{equation} \label{l3.2}
	\begin{array}{rl}
    |\Sigma^\text{\tiny{{PITC}}}_{\langle x, i\rangle \langle x', j\rangle }| \hspace{-2.8mm}& \displaystyle= |\Gamma_{\langle x, i\rangle \langle x', j\rangle}| \\
	&\displaystyle= |\sigma_{\langle x, i\rangle \langle x', j\rangle } - \Sigma_{\langle x, i\rangle \langle x', j\rangle |U}| \\
	&\displaystyle\leq |\sigma_{\langle x, i\rangle \langle x', j\rangle }|+|\Sigma_{\langle x, i\rangle \langle x', j\rangle |U}| \\
	&\displaystyle\leq 2|\sigma_{\langle x, i\rangle \langle x', j\rangle }|\ . 
	\end{array}
	\end{equation}
The first equality is due to \eqref{PITC.var} while the second equality follows from the definition of $\Gamma_{\langle x, i\rangle \langle x', j\rangle}$ (see last paragraph of Section~\ref{PITC}).
	 The last inequality follows from the practical assumption of absence of suppressor variables \cite{Das2008}: $|\Sigma_{\langle x, i\rangle \langle x', j\rangle |U}|\leq|\sigma_{\langle x, i\rangle \langle x', j\rangle}|$.
	 % in many practical cases, further conditioning does not make ... more correlated.
	%which has been shown to be true in many real world problems \cite{Das2008}.
	Then, 
	$$
	|\Sigma^\text{\tiny{{PITC}}}_{\langle x, i\rangle \langle x', j\rangle|X} | \leq |\Sigma^\text{\tiny{{PITC}}}_{\langle x, i\rangle \langle x', j\rangle }| \leq 2|\sigma_{\langle x, i\rangle \langle x', j\rangle }|\ .
	$$
	The first inequality follows from the practical assumption of absence of suppressor variables \cite{Das2008}. The second inequality is due to \eqref{l3.1} and \eqref{l3.2}.$\quad_\Box$\\

%\subsection{Proof of Lemma \ref{lemma.g}}
\noindent
\emph{Main Proof}. 
Let $B \triangleq  V_t \setminus X_t$. Using the spectral theorem, $(\Sigma^\text{\tiny{{PITC}}}_{BB|X})^{-1} = WQW^\top$ where the columns of $W$ are the eigenvectors of $(\Sigma^\text{\tiny{{PITC}}}_{BB|X})^{-1}$ and $Q$ is a diagonal matrix comprising the eigenvalues of $(\Sigma^\text{\tiny{{PITC}}}_{BB|X})^{-1}$. Let $\lambda_{\text{max}}(A)$ and $\lambda_{\text{min}}(A)$ denote, respectively, the maximum and minimum eigenvalues of matrix $A$, and $\alpha \triangleq W^\top\Sigma^\text{\tiny{{PITC}}}_{B\langle x, i\rangle |X}$. 
\begin{equation}\label{lemma.g1}
\hspace{-1.8mm}
\begin{array}{l}
\Sigma^\text{\tiny{{PITC}}}_{\langle x, i\rangle\langle x, i\rangle |X} - \Sigma^\text{\tiny{{PITC}}}_{\langle x, i\rangle\langle x, i\rangle |X\cup V_t \setminus X_t} \vspace{1mm}\\
\displaystyle= \Sigma^\text{\tiny{{PITC}}}_{\langle x, i\rangle\langle x, i\rangle |X}\ - \\\displaystyle\quad\left(\Sigma^\text{\tiny{{PITC}}}_{\langle x, i\rangle\langle x, i\rangle |X} - \Sigma^\text{\tiny{{PITC}}}_{\langle x, i\rangle B|X} (\Sigma^\text{\tiny{{PITC}}}_{BB|X})^{-1} \Sigma^\text{\tiny{{PITC}}}_{B\langle x, i\rangle |X}\right) \\
\displaystyle= \Sigma^\text{\tiny{{PITC}}}_{\langle x, i\rangle B|X} (\Sigma^\text{\tiny{{PITC}}}_{BB|X})^{-1} \Sigma^\text{\tiny{{PITC}}}_{B\langle x, i\rangle |X} \\
\displaystyle= \Sigma^\text{\tiny{{PITC}}}_{\langle x, i\rangle B|X} WQW^\top \Sigma^\text{\tiny{{PITC}}}_{B\langle x, i\rangle |X} \\
\displaystyle= \alpha^\top Q\alpha \\
\displaystyle \leq \lambda_{\text{max}}((\Sigma^\text{\tiny{{PITC}}}_{BB|X})^{-1}) \alpha^\top \alpha \\
\displaystyle= \frac{\Sigma^\text{\tiny{{PITC}}}_{\langle x, i\rangle B|X} WW^\top \Sigma^\text{\tiny{{PITC}}}_{B\langle x, i\rangle |X}}{\lambda_{\text{min}}(\Sigma^\text{\tiny{{PITC}}}_{BB|X})} \\
\displaystyle= \frac{\lVert \Sigma^\text{\tiny{{PITC}}}_{\langle x, i\rangle B|X} \rVert_2^2}{\lambda_{\text{min}}(\Sigma^\text{\tiny{{PITC}}}_{BB|X})} \\
\displaystyle= \frac{\sum_{\langle x', t\rangle \in B} |\Sigma^\text{\tiny{{PITC}}}_{\langle x, i\rangle  \langle x', t\rangle |X}|^2}{\lambda_{\text{min}}(\Sigma^\text{\tiny{{PITC}}}_{BB|X})} \\
\displaystyle\leq \frac{\sum_{\langle x', t\rangle \in B} 4|\sigma_{\langle x, i\rangle \langle x', t\rangle}|^2}{\lambda_{\text{min}}(\Sigma^\text{\tiny{{PITC}}}_{BB|X})} \vspace{1mm}\\
\displaystyle\leq \frac{4\sigma^2_{s_i} \sigma^2_{s_t} \sum_{\langle x', t\rangle \in B}\mathcal{N}( x -  x'| \underline{0}, P_0^{-1}+P^{-1}_i+P^{-1}_t)^2} {\sigma^2_{n_t}} \vspace{1mm}\\
%\displaystyle\leq \frac{4|V_t| |\sigma_{\langle x, i\rangle\langle x, t\rangle}|^2} {\sigma^2_{n_t}} \vspace{1mm}\\
\displaystyle
%= \frac{4|V_t| \sigma^2_{s_i} \sigma^2_{s_t}}{\sigma^2_{n_t}} = 4|V_t|\rho_t\sigma^2_{s_i}
= 4\rho_t\sigma^2_{s_i}R(\langle x, i\rangle,B)\ .
\end{array}
\end{equation}
%The fifth equality is due to the fact that $\lVert A \rVert_2 = \lambda_{max}(A)$ if $A$ is a symmetric, positive definite matrix as shown in Section 10.4.5 of \cite{petersen2012}.
The first equality is due to the incremental update formula of GP posterior variance (see Appendix C in \cite{Xu2014}).
The first inequality is due to the fact that $Q$ is a diagonal matrix comprising the eigenvalues of $(\Sigma^\text{\tiny{{PITC}}}_{BB|X})^{-1}$. 
The fifth equality is due to a property of eigenvalues that $\lambda_{\text{max}}(A^{-1}) = 1/\lambda_{\text{min}}(A)$. 
The sixth equality follows from the fact that $WW^\top = I$. 
The second inequality follows from Lemma~\ref{lemma0}. 
The third inequality is due to \eqref{kernel} and the fact that 
%$\sigma_{\langle x, i\rangle \langle x', t\rangle}\leq \sigma_{\langle x, i\rangle\langle x, t\rangle}$ \eqref{kernel} and 
$\lambda_{\text{min}}(\Sigma^\text{\tiny{{PITC}}}_{BB|X}) = \lambda_{\text{min}}(\Sigma^\text{\tiny{{PITC}}}_{BB|X}-\sigma^2_{n_t} I+\sigma^2_{n_t} I)=\lambda_{\text{min}}(\Sigma^\text{\tiny{{PITC}}}_{BB|X}-\sigma^2_{n_t} I) + \sigma^2_{n_t}\geq \sigma^2_{n_t}$ since $\lambda_{\text{min}}(\Sigma^\text{\tiny{{PITC}}}_{BB|X}-\sigma^2_{n_t} I)\geq 0$ (i.e., $\Sigma^\text{\tiny{{PITC}}}_{BB|X}-\sigma^2_{n_t} I$ is a positive semi-definite matrix).
%The second last inequality is due to \eqref{kernel}. 
Then,
$$
\begin{array}{l}
H(Y_{\langle x, i\rangle }|Y_X)-H(Y_{\langle x, i\rangle }|Y_{X\cup V_t \setminus X_t})\vspace{1mm} \\
\displaystyle= \frac{1}{2} \log \frac{\Sigma^\text{\tiny{PITC}}_{\langle x, i\rangle\langle x, i\rangle|X}}{\Sigma^\text{\tiny{PITC}}_{\langle x, i\rangle\langle x, i\rangle|X\cup B}} \vspace{1mm}\\
\displaystyle\leq \frac{1}{2} \log \frac{\Sigma^\text{\tiny{PITC}}_{\langle x, i\rangle\langle x, i\rangle|X\cup B} + 4\rho_t\sigma^2_{s_i}R(\langle x, i\rangle,B)}{\Sigma^\text{\tiny{PITC}}_{\langle x, i\rangle\langle x, i\rangle|X\cup B}} \vspace{1mm}\\
\displaystyle\leq \frac{1}{2} \log\left(1+\frac{4\rho_t\sigma^2_{s_i}R(\langle x, i\rangle,B)}{\sigma^2_{n_i}}\right) \\
\displaystyle= \frac{1}{2} \log (1+4\rho_t\rho_i R(\langle x, i\rangle,B))\ .
\end{array}
$$
The first inequality is due to \eqref{lemma.g1} while the second inequality follows from Lemma~\ref{lemma1}.

%Similar, we can get a lower bound as follows:
%
%\begin{align}
%& \quad \sigma^2_{ x_s|X} - \sigma^2_{ x_s|X, V_t \setminus X_t} \notag\\
%&= \alpha^\topQ\alpha \notag\\
%& \geq \lambda_{min}((\Sigma^{PITC}_{BB|X})^{-1}) \alpha^\top \alpha \notag\\
%&= \frac{\sum_{ x \in B} |\sigma^{PITC}_{ x_s  x|X}|^2}{\lambda^{max}(\Sigma^{PITC}_{BB|X})} \notag\\
%&\geq \frac{2b \sigma^2_{n_x}} {\sigma^2_{n_t}} \notag\\
%&\leq \frac{2b \sigma^2_{s_s} \sigma^2_{s_t}}{\sigma^2_{n_t}} = 2b\rho_t\sigma^2_{s_s}, \label{lemma.g2}
%\end{align}
%
\section{Proof of Theorem~\ref{thm:time}} \label{a2.5}
%The second term of \eqref{h3} can be written as
If $i=t$, then
$$H(Y_{\langle x, t\rangle}|Y_X)= \frac{1}{2} \log (2\pi e)\Sigma^\text{\tiny{PITC}}_{\langle x, t\rangle\langle x, t\rangle|X}$$ where
$\Sigma^\text{\tiny{PITC}}_{\langle x, t\rangle\langle x, t\rangle|X}$ is previously defined in \eqref{PITC.var}. So, evaluating $H(Y_{\langle x, t\rangle}|Y_X)$ incurs $\mathcal{O}(|U|^2)$ time for every $\langle x, t\rangle\in V_t\setminus X_t$ and $\mathcal{O}(|U|^3+N^3)$ time in each iteration; this worst-case time complexity occurs when all the tuples in $X$ are of one measurement type.

Otherwise (i.e., $i \neq t$),
$$
\begin{array}{l}
\displaystyle H(Y_{\langle x, i\rangle}|Y_X)- H(Y_{\langle x, i\rangle}|Y_{X\cup V_t \setminus X_t})  \vspace{1mm}\\
=\displaystyle \frac{1}{2} \log \frac{\Sigma^\text{\tiny{PITC}}_{\langle x, i\rangle\langle x, i\rangle|X}}{\Sigma^\text{\tiny{PITC}}_{\langle x, i\rangle\langle x, i\rangle|X\cup V_t \setminus X_t}} \vspace{1mm}\\
=\displaystyle \frac{1}{2} \log\frac{\Sigma^\text{\tiny{PITC}}_{\langle x, i\rangle\langle x, i\rangle|X}}{\Sigma^\text{\tiny{PITC}}_{\langle x, i\rangle\langle x, i\rangle |\bigcup_{i \neq t} X_i\cup V_t}}
\end{array}
$$
where
$\Sigma^\text{\tiny{PITC}}_{\langle x, i\rangle\langle x, i\rangle|X}$ and $\Sigma^\text{\tiny{PITC}}_{\langle x, i\rangle\langle x, i\rangle |\bigcup_{i \neq t} X_i\cup V_t}$ are previously defined in \eqref{PITC.var}. Therefore, evaluating $\Sigma^\text{\tiny{PITC}}_{\langle x, i\rangle\langle x, i\rangle|X}$ incurs $\mathcal{O}(|U|^2)$ time for every $\langle x, i\rangle\in V_{\text{-}t} \setminus X_{\text{-}t}$ and $\mathcal{O}(|U|^3+N^3)$ time in each iteration; this worst-case time complexity occurs when all the tuples in $X$ are of one measurement type.

Let $A \triangleq \bigcup_{i \neq t} X_i \cup V_t$. 
%The time complexity of evaluating $|\Sigma_{UU|V_t, \bigcup_{i \neq t} X_i}| = |\Sigma_{UU|B}| $ is dominated by a $|V_t|^3$ term that is incurred due to the inversion of $\Lambda_B$. However, 
Then, by the definition of $\Lambda_{A}$ (see last paragraph of Section~\ref{PITC}),
$$
\Sigma_{UA} \Lambda^{-1}_{A} \Sigma_{AU} = \sum_{i \neq t} \Sigma_{UX_i} \Sigma_{X_i X_i|U}^{-1} \Sigma_{X_iU}+\Sigma_{UV_t} \Sigma_{V_tV_t|U}^{-1} \Sigma_{V_tU}\ .
$$
Evaluating the $\sum_{i \neq t} \Sigma_{UX_i} \Sigma_{X_i X_i|U}^{-1} \Sigma_{X_iU}$ term incurs $\mathcal{O}(|U|^3+N^3)$ time in each iteration; this worst-case time complexity occurs when all the tuples in $X$ are of one measurement type.
Note that the $\Sigma_{UV_t} \Sigma_{V_tV_t|U}^{-1} \Sigma_{V_tU}$ term remains the same in each iteration (i.e., since it is independent of $X$) and hence only needs to be computed once in $\mathcal{O}(|V_t|^3)$ time in our approximation algorithm.
As a result, evaluating $\Sigma^\text{\tiny{PITC}}_{\langle x, i\rangle\langle x, i\rangle |\bigcup_{i \neq t} X_i\cup V_t}= \Sigma^\text{\tiny{PITC}}_{\langle x, i\rangle\langle x, i\rangle |A}$ (specifically, its efficient formulation exploiting $\Sigma_{UA} \Lambda^{-1}_{A} \Sigma_{AU}$, as shown in \cite{Alvarez2011}) incurs $\mathcal{O}(|U|^2)$ time for every $\langle x, i\rangle\in V_{\text{-}t} \setminus X_{\text{-}t}$ and $\mathcal{O}(|U|^3+N^3)$ time in each iteration, and a \emph{one-off} cost of 
$\mathcal{O}(|V_t|^3)$ time.
Consequently, evaluating $H(Y_{\langle x, i\rangle}|Y_X)- H(Y_{\langle x, i\rangle}|Y_{X\cup V_t \setminus X_t})$ incurs $\mathcal{O}(|U|^2)$ time for every $\langle x, i\rangle\in V_{\text{-}t} \setminus X_{\text{-}t}$ and $\mathcal{O}(|U|^3+N^3)$ time in each iteration, and a \emph{one-off} cost of $\mathcal{O}(|V_t|^3)$ time.

Since $|U|\leq |V_t|<|V|$, our approximation algorithm thus incurs $\mathcal{O}(N(|V||U|^2+N^3)+|V_t|^3)$ time.
\section{Proof of Lemma~\ref{near-s}}\label{a3}
To prove that $F(X)$ is $\epsilon$-submodular, we have to show that
$$
F(X' \cup \{\langle x, i\rangle\} ) - F(X') \leq F(X \cup \{\langle x, i\rangle\} ) - F(X) + \epsilon
$$
for any $X \subseteq X' \subseteq V$ and $\langle x, i\rangle  \in V \setminus X'$. Before doing this, the following lemma is needed:
\begin{lemma}\label{lemma2}
	 Suppose that $\epsilon_1\geq 0$ is given. For any $\langle x, i\rangle  \in V_{\text{-}t} \setminus X'_{\text{-}t}$, if $\Sigma^\text{\tiny{\emph{PITC}}}_{\langle x, i\rangle\langle x, i\rangle  | X\cup V_t \setminus X'_t} - \Sigma^\text{\tiny{\emph{PITC}}}_{\langle x, i\rangle\langle x, i\rangle  | X'\cup V_t \setminus X'_t} \leq \epsilon_1$, then $I(Y_{\langle x, i\rangle };Y_{V_t \setminus X'_t}|Y_{X'}) \leq I(Y_{\langle x, i\rangle };Y_{V_t \setminus X'_t}|Y_X) + \epsilon$ where $\epsilon = 0.5 \log(1+ \epsilon_1/\sigma^2_{n^*})$.
\end{lemma}

\noindent\emph{Proof}. Let $\bar{X}\triangleq X'\setminus X$.
%\begin{equation}\label{eq1}
%	\begin{array}{l}
%\displaystyle	\Sigma^\text{\tiny{{PITC}}}_{\langle x, i\rangle\langle x, i\rangle  | X\cup V_t \setminus X'_t} - \Sigma^\text{\tiny{{PITC}}}_{\langle x, i\rangle\langle x, i\rangle  |  \bar{X}\cup X\cup V_t \setminus X'_t}\vspace{1mm} \\
%\displaystyle	\leq \Sigma^\text{\tiny{{PITC}}}_{\langle x, i\rangle\langle x, i\rangle  | V_t \setminus X'_t} - \Sigma^\text{\tiny{{PITC}}}_{\langle x, i\rangle\langle x, i\rangle  | \bar{X}\cup V_t \setminus X'_t} \\
%\displaystyle	\leq \epsilon_1 \ .
%	\end{array}
%\end{equation}	
%	The first inequality follows from the property that variance reduction $G(A)\triangleq \Sigma^\text{\tiny{{PITC}}}_{\langle x, i\rangle\langle x, i\rangle} - \Sigma^\text{\tiny{{PITC}}}_{\langle x, i\rangle\langle x, i\rangle|A}$ is submodular in many practical cases \cite{Das2008} while the second inequality is due to the sufficient condition.
	Then,
	\begin{equation}\label{eq2}
	\begin{array}{l}
\displaystyle	I(Y_{\langle x, i\rangle }; Y_{\bar{X}} | Y_{X\cup V_t \setminus X'_t}) \vspace{1mm}\\
\displaystyle= H(Y_{\langle x, i\rangle } | Y_{X\cup V_t \setminus X'_t}) - H(Y_{\langle x, i\rangle } | Y_{\bar{X}\cup X\cup V_t \setminus X'_t}) \vspace{1mm}\\
\displaystyle= \frac{1}{2} \log \frac{\Sigma^\text{\tiny{{PITC}}}_{\langle x, i\rangle\langle x, i\rangle  | X\cup V_t \setminus X'_t}}{\Sigma^\text{\tiny{{PITC}}}_{\langle x, i\rangle\langle x, i\rangle  |  \bar{X}\cup X\cup V_t \setminus X'_t}} \vspace{1mm}\\
\displaystyle\leq \frac{1}{2} \log \frac{\Sigma^\text{\tiny{{PITC}}}_{\langle x, i\rangle\langle x, i\rangle  |  \bar{X}\cup X\cup V_t \setminus X'_t} + \epsilon_1} {\Sigma^\text{\tiny{{PITC}}}_{\langle x, i\rangle\langle x, i\rangle  |  \bar{X}\cup X\cup V_t \setminus X'_t}} \vspace{1mm}\\
\displaystyle= \frac{1}{2} \log \left(1 + \frac {\epsilon_1}{\Sigma^\text{\tiny{{PITC}}}_{\langle x, i\rangle\langle x, i\rangle  |  \bar{X}\cup X\cup V_t \setminus X'_t}}\right) \vspace{1mm}\\
\displaystyle\leq \frac{1}{2} \log \left(1 + \frac {\epsilon_1} {\sigma^2_{n_i}}\right)\vspace{1mm} \\
\displaystyle\leq \frac{1}{2} \log \left(1 + \frac {\epsilon_1} {\sigma^2_{n^*}}\right). 
	\end{array}
	\end{equation}
%	The first inequality is due to \eqref{eq1}. 
The first inequality is due to the sufficient condition. 
	The second inequality follows from Lemma~\ref{lemma1}.
	Then, by the definition of conditional mutual information, 
	$$
	\begin{array}{l}
	\displaystyle I(Y_{\langle x, i\rangle }; Y_{V_t \setminus X'_t} | Y_{\bar{X}\cup X}) + I(Y_{\langle x, i\rangle }; Y_{\bar{X}} | Y_X) \vspace{1mm}\\
	\displaystyle= H(Y_{\langle x, i\rangle }| Y_{\bar{X}\cup X}) - H(Y_{\langle x, i\rangle } | Y_{\bar{X}\cup X\cup V_t \setminus X'_t})\ + \\
	\displaystyle \quad H(Y_{\langle x, i\rangle } | Y_X) - H(Y_{\langle x, i\rangle } |Y_{\bar{X}\cup X})\vspace{1mm} \\
	\displaystyle= H(Y_{\langle x, i\rangle } | Y_X) - H(Y_{\langle x, i\rangle } | Y_{\bar{X}\cup X\cup V_t \setminus X'_t})\vspace{1mm} \\
	\displaystyle= H(Y_{\langle x, i\rangle } | Y_X) - H(Y_{\langle x, i\rangle } | Y_{X\cup V_t \setminus X'_t}) \ + \\
	\displaystyle \quad H(Y_{\langle x, i\rangle } | Y_{X\cup V_t \setminus X'_t}) - H(Y_{\langle x, i\rangle } | Y_{\bar{X}\cup X\cup V_t \setminus X'_t}) \vspace{1mm}\\
	\displaystyle= I(Y_{\langle x, i\rangle }; Y_{V_t \setminus X'_t} | Y_X) + I(Y_{\langle x, i\rangle }; Y_{\bar{X}} | Y_{X\cup V_t \setminus X'_t}) \ .
	\end{array}
	$$
	Therefore,
	$$
	\begin{array}{l}
	\displaystyle I(Y_{\langle x, i\rangle }; Y_{V_t \setminus X'_t} | Y_{X'}) \vspace{1mm}\\
	\displaystyle = I(Y_{\langle x, i\rangle }; Y_{V_t \setminus X'_t} | Y_{\bar{X}\cup X}) \vspace{1mm}\\
	\displaystyle = I(Y_{\langle x, i\rangle }; Y_{V_t \setminus X'_t} | Y_X) + I(Y_{\langle x, i\rangle }; Y_{\bar{X}} | Y_{X\cup V_t \setminus X'_t})\ - \\
	\displaystyle  \quad I(Y_{\langle x, i\rangle }; Y_{\bar{X}} | Y_X) \vspace{1mm}\\
	\displaystyle  \leq I(Y_{\langle x, i\rangle }; Y_{V_t \setminus X'_t} | Y_X) + I(Y_{\langle x, i\rangle }; Y_{\bar{X}} | Y_{X\cup V_t \setminus X'_t})\\
	\displaystyle  \leq I(Y_{\langle x, i\rangle }; Y_{V_t \setminus X'_t} | Y_X) + 0.5\log \left(1 + \frac {\epsilon_1} {\sigma^2_{n^*}}\right) \ .
	\end{array}
	$$
	The first inequality is due to the fact that conditional mutual information is non-negative. The last inequality follows from \eqref{eq2}.$\quad_\Box$\\

\noindent
\emph{Main Proof}.
%\subsection{Proof of Theorem \ref{near-s}}
To prove that $F(X)$ is $\epsilon$-submodular, we have to show that
$
H(Y_{\langle x, i\rangle }|Y_{X'}) - \delta_i H(Y_{\langle x, i\rangle }|Y_{X'\cup V_t \setminus X_t'}) \leq 
H(Y_{\langle x, i\rangle }|Y_X) - \delta_i H(Y_{\langle x, i\rangle }|Y_{X\cup V_t \setminus X_t}) + \epsilon
$
for any $X \subseteq X' \subseteq V$ and $\langle x, i\rangle  \in V \setminus X'$.

If $i=t$, then $H(Y_{\langle x, i\rangle }|Y_{X'}) \leq H(Y_{\langle x, i\rangle }|Y_X) \leq H(Y_{\langle x, i\rangle }|Y_X) + \epsilon$ for any $\epsilon\geq 0$ due to the ``information never hurts'' bound for entropy \cite{Cover91}. 

Otherwise (i.e., $i\neq t$),
$$
\begin{array}{l}
\displaystyle H(Y_{\langle x, i\rangle }|Y_{X'}) - H(Y_{\langle x, i\rangle }|Y_{X'\cup V_t \setminus X_t'}) \\
\displaystyle \leq H(Y_{\langle x, i\rangle }|Y_X) - H(Y_{\langle x, i\rangle }|Y_{X\cup V_t \setminus X_t'}) + \epsilon \\
\displaystyle \leq H(Y_{\langle x, i\rangle }|Y_X) - H(Y_{\langle x, i\rangle }|Y_{X\cup V_t \setminus X_t}) + \epsilon
\end{array}
$$
where $\epsilon = 0.5\log \left(1 + {\epsilon_1}/{\sigma^2_{n^*}}\right)$. The first inequality is due to Lemma~\ref{lemma2}. The second inequality follows from the ``information never hurts'' bound for entropy \cite{Cover91}: $H(Y_{\langle x, i\rangle }|Y_{X\cup V_t \setminus X_t'}) \geq H(Y_{\langle x, i\rangle }|Y_{X\cup V_t \setminus X_t})$ since $(V_t \setminus X'_t) \subseteq (V_t \setminus X_t)$.
\section{Proof of Theorem~\ref{bound}} \label{a5}
Our proof here is similar to that of Theorem $1.5$ in \cite{Krause2012} which is a generalization of the well-known result of \citeauthor{Nemhauser1978}~\shortcite{Nemhauser1978}. 
The key difference is that we exploit $\epsilon$-submodularity of $F(X)$ (i.e., Lemma~\ref{near-s}) instead of submodularity, as shown below for completeness.

%we replace the inequalities following from submodularity in the proof of Theorem $1.5$ in \cite{Krause2012}  with that due to %We only need to change the submodulary inequality with the near submodular one as shown below.

Let $X^* \triangleq \{ \langle x_1, s_1\rangle^*, \ldots,  \langle x_N, s_N\rangle^*\}$ be the optimal set of selected observations, $X^k$ be the set of tuples selected by our approximation algorithm in iteration $k = 1,\ldots, N$, $X^0\triangleq \emptyset$, and $\Delta (\langle x, i\rangle |X) \triangleq F(X \cup \{\langle x, i\rangle\} ) - F(X)$ . Then, 
$$
\begin{array}{l}
 F(X^*)  \\
\displaystyle\leq  F(X^* \cup X^k)  \\
\displaystyle= F(X^k)+\sum_{j=1}^N \Delta\left(\langle x_j, s_j\rangle^* \Bigg| \bigcup_{r=1}^{j-1} \{\langle x_r, s_r\rangle^*\} \cup X^k\right)  \\
\displaystyle\leq  F(X^k) + \sum_{j=1}^N \left(\Delta (\langle x_j, s_j\rangle^*| X^k) + \epsilon\right)  \\
\displaystyle\leq  F(X^k) + \sum_{j=1}^N \left(F(X^{k+1}) - F(X^k) + \epsilon\right)  \\
\displaystyle\leq F(X^k) + N \left(F(X^{k+1}) - F(X^k) + \epsilon\right).
\end{array}
$$
The first inequality follows from the nondecreasing property of $F(X)$. The first equality is a straightforward telescoping sum. The second inequality follows from the $\epsilon$-submodularity of $F(X)$, as proven in Lemma~\ref{near-s}. The third inequality follows from \eqref{greedy}. Then,  
\begin{equation} \label{a5.1}
F(X^*) - F(X^k) \leq N \left(F(X^{k+1}) - F(X^k) + \epsilon\right).
\end{equation}
Let $\zeta_k \triangleq F(X^*) - F(X^k)$. Then, \eqref{a5.1} can be rewritten as $\zeta_k \leq N(\zeta_k-\zeta_{k+1}+\epsilon)$ which can be rearranged to yield
\begin{equation} \label{a5.2}
\zeta_{k+1} \leq \left( 1-\frac{1}{N}\right) \zeta_k + \epsilon\ .
\end{equation}
Then, by recursion of \eqref{a5.2}, it is straightforward to get
\begin{equation} \label{a5.3}
\zeta_k \leq \left( 1-\frac{1}{N}\right)^k \zeta_0 + N\left( 1- \left(1-\frac{1}{N}\right)^k\right)\epsilon \ .
\end{equation}
Then, by substituting $\zeta_k = F(X^*)-F(X^k)$ and $\zeta_0 = F(X^*)-F(X^0) = F(X^*)$,  \eqref{a5.3} can be rearranged to
$$
\begin{array}{rl}
F(X^k)\hspace{-2.8mm} &\displaystyle\geq \left(1-\left( 1-\frac{1}{N}\right)^k \right)(F(X^*)-N\epsilon) \\
&\displaystyle\geq (1-e^{-k/N})(F(X^*)-N\epsilon)\ .
\end{array}
$$
The second inequality follows from the well-known inequality $e^{-x}\geq 1-x$. Finally, Theorem~\ref{bound} is obtained when $k=N$ and $\epsilon = 0.5 \log(1+ \epsilon_1 / \sigma^2_{n^*})$, as defined in Lemma~\ref{near-s}.
\section{Proof of Lemma~\ref{lemma}}\label{a4}
Let $B \triangleq \widetilde{X}\cup V_t \setminus X_t$ and $A\triangleq X\setminus \widetilde{X}$. From the incremental update formula of GP posterior variance (see Appendix C in \cite{Xu2014}), 
\begin{equation}\label{D.0}
\begin{array}{l}
\displaystyle\Sigma^\text{\tiny{{PITC}}}_{\langle x, i\rangle\langle x, i\rangle  | B} - \Sigma^\text{\tiny{{PITC}}}_{\langle x, i\rangle\langle x, i\rangle  | B\cup A}\vspace{1mm}\\
\displaystyle= \Sigma^\text{\tiny{{PITC}}}_{\langle x, i\rangle\langle x, i\rangle  | B}\ -\\ \displaystyle\quad\left(\Sigma^\text{\tiny{{PITC}}}_{\langle x, i\rangle\langle x, i\rangle  | B} - \Sigma^\text{\tiny{{PITC}}}_{\langle x, i\rangle A|B} (\Sigma^\text{\tiny{{PITC}}}_{AA|B})^{-1} \Sigma^\text{\tiny{{PITC}}}_{A\langle x, i\rangle |B}\right) \vspace{1mm}\\
\displaystyle= \Sigma^\text{\tiny{{PITC}}}_{\langle x, i\rangle A|B} (\Sigma^\text{\tiny{{PITC}}}_{AA|B})^{-1} \Sigma^\text{\tiny{{PITC}}}_{A\langle x, i\rangle |B}\ . 
\end{array}
\end{equation}
Let $\Sigma^\text{\tiny{{PITC}}}_{AA|B} \triangleq C+E$ where $C$ is defined as a matrix with the same diagonal components as $\Sigma^\text{\tiny{{PITC}}}_{AA|B}$ and off-diagonal components 0 while $E$ is defined as a matrix with diagonal components $0$ and the same off-diagonal components as $\Sigma^\text{\tiny{{PITC}}}_{AA|B}$. Then,
\begin{equation}\label{D.1}
\begin{array}{rl}
\lVert C^{-1} \rVert_2 \hspace{-2.8mm}&=\displaystyle \lambda_\text{max}(C^{-1}) \\
&=\displaystyle \frac{1}{\lambda_\text{min}(C)} \\
&=\displaystyle \frac{1}{\min_{\langle x, i\rangle \in A} \Sigma^\text{\tiny{{PITC}}}_{\langle x, i\rangle\langle x, i\rangle  | B}} \\
&\leq \displaystyle\frac{1}{\sigma^2_{n_i}} \leq \frac{1}{\sigma^2_{n^*}} \ .
\end{array}
\end{equation}
The first equality is due to a property of matrix norm in Section $10.4.5$ in \cite{petersen2012}. 
The second equality is due to a property of eigenvalues that $\lambda_{\text{max}}(C^{-1}) = 1/\lambda_{\text{min}}(C)$. 
The third equality is due to the diagonal property of $C$. The first inequality is due to Lemma~\ref{lemma1}.

Matrix $E$ comprises off-diagonal components $\Sigma^\text{\tiny{{PITC}}}_{\langle x, i\rangle\langle x', j\rangle |B}$ for all $\langle x, i\rangle, \langle x', j\rangle  \in A$ such that $\langle x, i\rangle\neq \langle x', j\rangle$, each of which has an absolute value not more than $2\sigma_{s^*}^2\xi^{p^2}$:
$$
\begin{array}{l}
\displaystyle |\Sigma^\text{\tiny{{PITC}}}_{\langle x, i\rangle\langle x', j\rangle |B}| \\
\displaystyle \leq 2|\sigma_{\langle x, i\rangle\langle x', j\rangle }| \\
\displaystyle= 2|\sigma_{s_i}\sigma_{s_j}| \mathcal{N}( x -  x'| \underline{0}, P_0^{-1}+P^{-1}_i+P^{-1}_j) \\
\displaystyle= 2|\sigma_{s_i}\sigma_{s_j}| \exp\left\{-\frac{1}{2} \sum_{v=1}^{d} \frac{(x_v-x'_v)^2}{\ell^{ij}_v} \right\} \vspace{1mm}\\
\displaystyle \leq 2|\sigma_{s_i}\sigma_{s_j}| \exp\left\{-\frac{(x_1-x'_1)^2}{2\ell^{ij}_1} \right\} \\
\displaystyle \leq 2|\sigma_{s_i}\sigma_{s_j}| \exp\left\{-\frac{p^2\omega^2}{2\ell}\right\} \\
\displaystyle= 2|\sigma_{s_i}\sigma_{s_j}| \xi^{p^2} \\
\displaystyle \leq 2\sigma^2_{s^*} \xi^{p^2}
\end{array}
$$
where $x_v$ is the $v$-th component of a $d$-dimensional location vector $x$ and $\ell^{ij}_v$ denotes the $v$-th diagonal component of $P_0^{-1}+P_i^{-1}+P_j^{-1}$.
The first inequality follows from Lemma~\ref{lemma0}. 
The second equality is due to the precision matrices being diagonal. 
The third inequality follows from $\ell \triangleq \max_{i,j\in\{1,\ldots,M\}} \ell^{ij}_1$ and the fact that the distance between $x_1$ and $x'_1$ of any $\langle x, i\rangle, \langle x', j\rangle  \in A$ must be at least $p\omega$ due to the construction of $V^-$.
Therefore,
\begin{equation} \label{D.2}
\lVert E \rVert_2 \leq 2N\sigma_{s^*}^2\xi^{p^2}
\end{equation}
due to a property that the $2$-norm of a matrix is at most its largest absolute component multiplied by its dimension \cite{Golub96}. 

Similarly, $\Sigma^\text{\tiny{{PITC}}}_{\langle x, i\rangle A|B}$ comprises components $\Sigma^\text{\tiny{{PITC}}}_{\langle x, i\rangle\langle x', j\rangle |B}$ for all $ \langle x', j\rangle  \in A$, each of which has an absolute value not more than $2\sigma_{s^*}^2\xi^{p^2}$:
%for any $\langle x', i\rangle \in A$ in matrix $\Sigma^{PITC}_{\langle x, i\rangle  A| B}$,
\begin{equation} \label{D.3}
|\Sigma^\text{\tiny{{PITC}}}_{\langle x, i\rangle\langle x', j\rangle |B}| \leq 2|\sigma_{\langle x, i\rangle \langle x', j\rangle}| \leq 2\sigma_{s^*}^2\xi^{p^2}.
\end{equation}
Now,
\begin{equation} \label{D.4}
\begin{array}{l}
\displaystyle\Sigma^\text{\tiny{{PITC}}}_{\langle x, i\rangle A|B} (C+E)^{-1} \Sigma^\text{\tiny{{PITC}}}_{A\langle x, i\rangle|B} - \Sigma^\text{\tiny{{PITC}}}_{\langle x, i\rangle A|B} C^{-1} \Sigma^\text{\tiny{{PITC}}}_{A\langle x, i\rangle|B}\vspace{1mm} \\
\displaystyle= \Sigma^\text{\tiny{{PITC}}}_{\langle x, i\rangle A|B} \left\{(C+E)^{-1} - C^{-1}\right\} \Sigma^\text{\tiny{{PITC}}}_{A\langle x, i\rangle|B} \vspace{1mm}\\
\displaystyle\leq \lVert \Sigma^\text{\tiny{{PITC}}}_{\langle x, i\rangle A|B}\rVert_2^2 \lVert(C+E)^{-1} - C^{-1}\rVert_2 \vspace{1mm} \\
\displaystyle \leq \sum_{\langle x', j\rangle \in A} |\Sigma^\text{\tiny{{PITC}}}_{\langle x, i\rangle\langle x', j\rangle |B}|^2 \frac{\lVert C^{-1} \rVert_2 \lVert E\rVert_2}{\frac{1}{\lVert C^{-1} \rVert_2} - \lVert E\rVert_2}  \\
\displaystyle\leq 4N\sigma_{s^*}^4\xi^{2p^2} \frac{\lVert C^{-1} \rVert_2 \lVert E\rVert_2}{\frac{1}{\lVert C^{-1} \rVert_2} - \lVert E\rVert_2}\ .
\end{array}
\end{equation}
The first inequality is due to Cauchy-Schwarz inequality and submultiplicativity of the matrix norm \cite{Stewart1990}. The second inequality follows from an important result in the perturbation theory of matrix inverses (in particular, Theorem III$.2.5$
in \cite{Stewart1990}). It requires the assumption $\lVert C^{-1}E\rVert_2 < 1$. Using \eqref{D.1}, \eqref{D.2}, and the matrix norm property in Section $10.4.2$ in \cite{petersen2012}, this assumption can be satisfied by 
$$
\lVert C^{-1}E\rVert_2 \leq \lVert C^{-1} \rVert_2 \lVert E\rVert_2 \leq \frac{2N\sigma_{s^*}^2\xi^{p^2}}{\sigma^2_{n^*}} < 1.
$$
Then,
\begin{equation}\label{D.5}
p^2 > \log\left(\frac{\sigma_{n^*}^2}{2N\sigma_{s^*}^2}\right) \Big/ \log \xi\ .
\end{equation}
The last inequality in \eqref{D.4} is due to \eqref{D.3}.
Then, from both \eqref{D.0} and \eqref{D.4},	
$$
\begin{array}{l}
\displaystyle\Sigma^\text{\tiny{{PITC}}}_{\langle x, i\rangle\langle x, i\rangle  | B} - \Sigma^\text{\tiny{{PITC}}}_{\langle x, i\rangle\langle x, i\rangle  | B\cup A}\vspace{1mm}\\
\displaystyle=\Sigma^\text{\tiny{{PITC}}}_{\langle x, i\rangle A|B} (C+E)^{-1} \Sigma^\text{\tiny{{PITC}}}_{A\langle x, i\rangle|B} \\
\displaystyle\leq \Sigma^\text{\tiny{{PITC}}}_{\langle x, i\rangle A|B} C^{-1} \Sigma^\text{\tiny{{PITC}}}_{A\langle x, i\rangle|B} + 4N\sigma_{s^*}^4\xi^{2p^2} \frac{\lVert C^{-1} \rVert_2 \lVert E\rVert_2}{\frac{1}{\lVert C^{-1} \rVert_2} - \lVert E\rVert_2} \vspace{1mm}\\
\displaystyle\leq \lVert \Sigma^\text{\tiny{{PITC}}}_{\langle x, i\rangle A|B} \rVert_2^2 \lVert C^{-1} \rVert_2 + 4N\sigma_{s^*}^4\xi^{2p^2} \frac{\lVert C^{-1} \rVert_2 \lVert E\rVert_2}{\frac{1}{\lVert C^{-1} \rVert_2} - \lVert E\rVert_2} \vspace{1mm}\\
\displaystyle\leq 4N\sigma_{s^*}^4\xi^{2p^2}\lVert C^{-1} \rVert_2 + 4N\sigma_{s^*}^4\xi^{2p^2} \frac{\lVert C^{-1} \rVert_2 \lVert E\rVert_2}{\frac{1}{\lVert C^{-1} \rVert_2} - \lVert E\rVert_2} \vspace{1mm}\\
\displaystyle= 4N\sigma_{s^*}^4\xi^{2p^2}\lVert C^{-1} \rVert_2 \left( 1+ \frac{\lVert E\rVert_2}{\frac{1}{\lVert C^{-1} \rVert_2} - \lVert E\rVert_2}\right) \\
\displaystyle= \frac{4N\sigma_{s^*}^4\xi^{2p^2}}{\frac{1}{\lVert C^{-1} \rVert_2} - \lVert E\rVert_2} \\
\displaystyle\leq \frac{ 4N\sigma_{s^*}^4\xi^{2p^2}}{\sigma^2_{n^*}-2N\sigma_{s^*}^2\xi^{p^2}}\ .
\end{array}
$$
The first inequality is due to \eqref{D.4}. The second inequality is due to Cauchy-Schwarz inequality. The third inequality is due to \eqref{D.3}. The last inequality follows from  \eqref{D.1} and \eqref{D.2}.

To satisfy \eqref{sert} in Lemma~\ref{near-s}, let 
$$\frac{ 4N\sigma_{s^*}^4\xi^{2p^2}}{\sigma^2_{n^*}-2N\sigma_{s^*}^2\xi^{p^2}} \leq \epsilon_1\ .$$
Then,
\begin{equation}\label{D.6}
p^2 \geq \log\left\{\frac{1}{4\sigma_{s^*}^2} \left( \sqrt{\epsilon_1^2+\frac{4\epsilon_1\sigma^2_{n^*}}{N}} - \epsilon_1 \right)  \right\} \Bigg/ \log \xi\ .
\end{equation}
Finally, from both \eqref{D.5} and \eqref{D.6}, Lemma~\ref{lemma} results.
\section{Jura and IEQ Datasets}\label{datasets}
\begin{figure}[h]
	\centering
	%\vspace{-3mm}
	\begin{tabular}{cc}
		\hspace{-6mm}\includegraphics[scale=0.32]{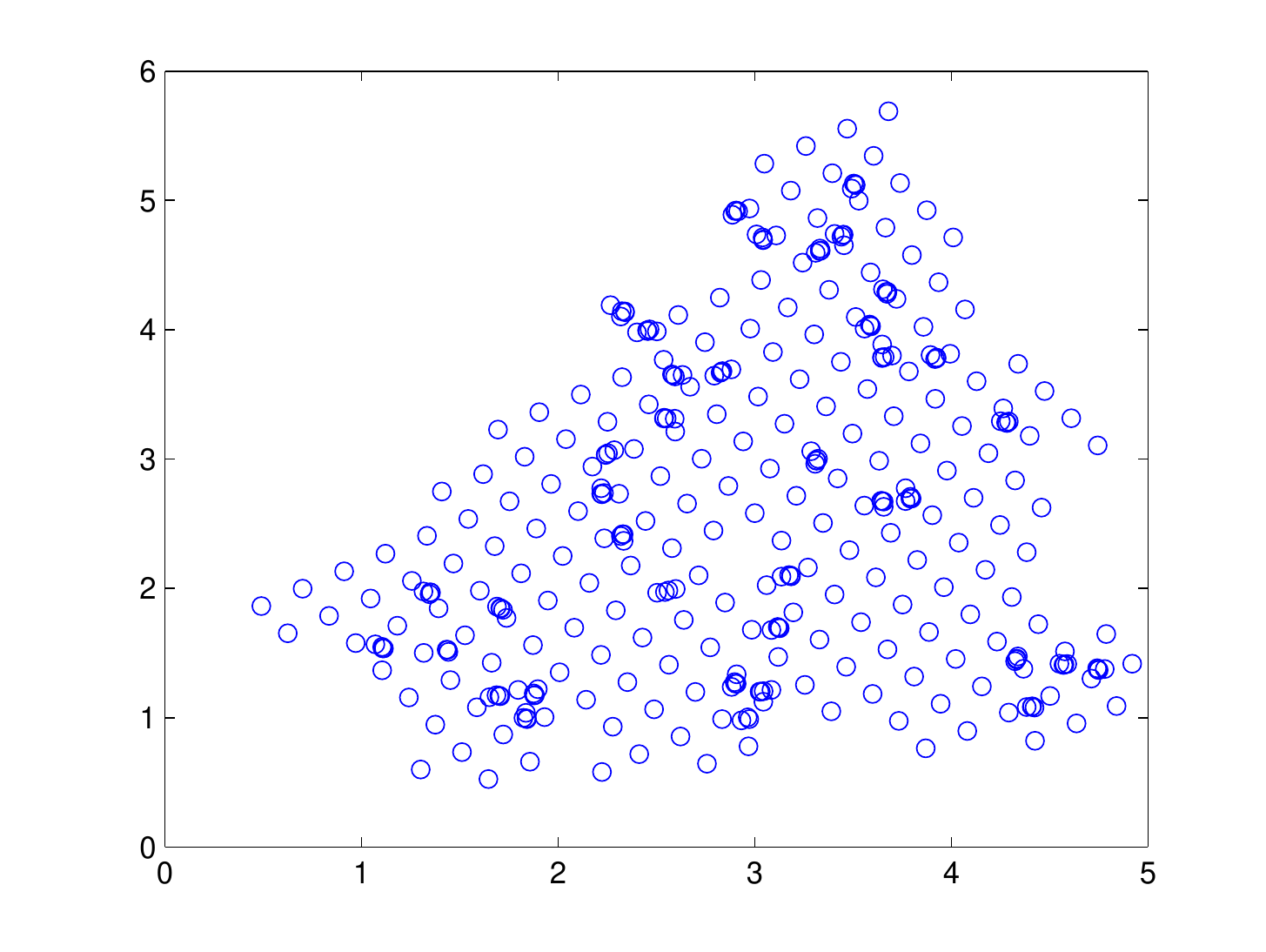} & \hspace{-8mm}\includegraphics[scale=0.32]{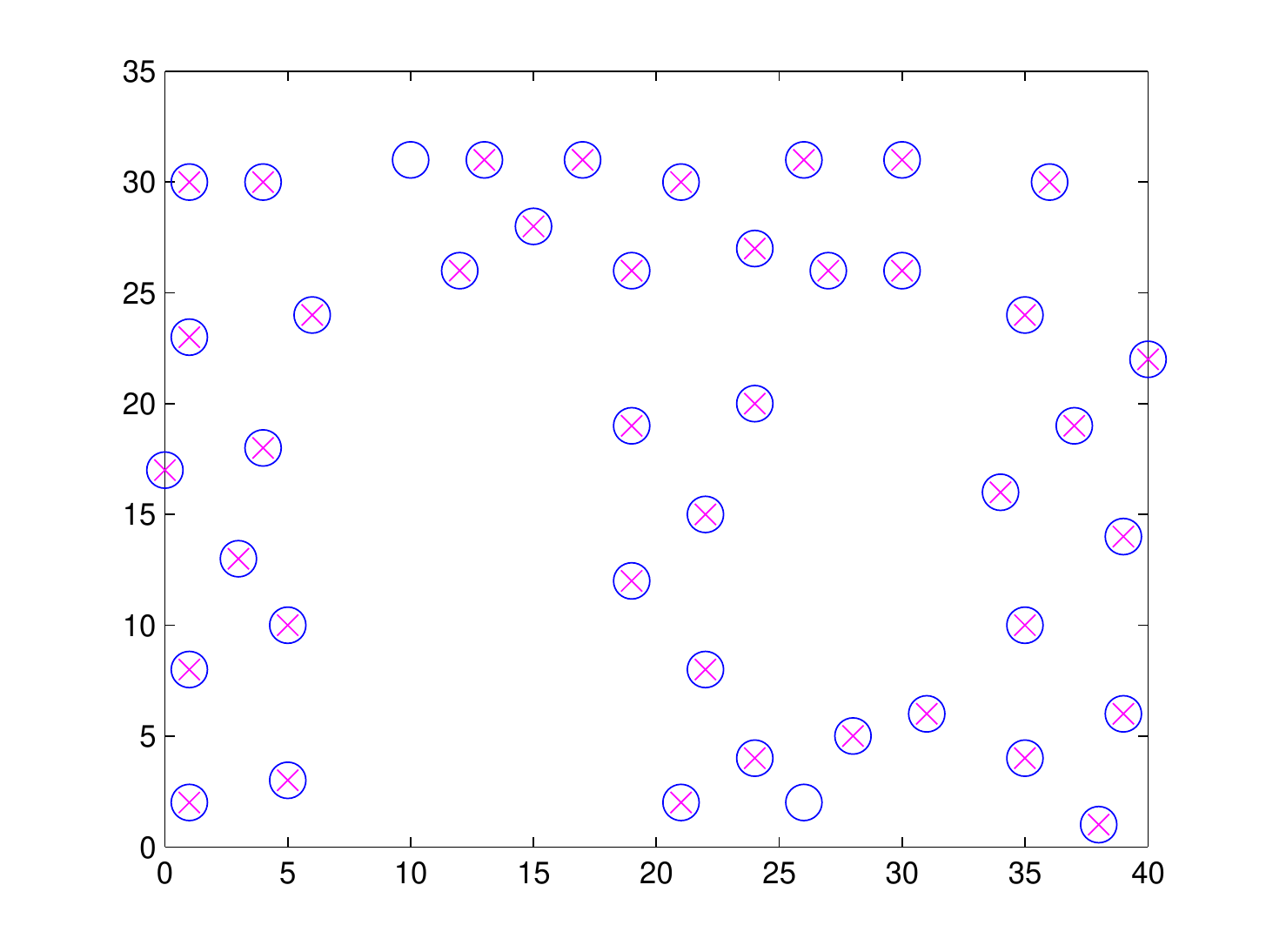} \vspace{-3mm}\\
		\hspace{-6mm}{(a)} & \hspace{-8mm}{(b)}
	\end{tabular}
	\caption{Sampling locations for the (a) Jura (km) and (b) IEQ (m) datasets where `$\circ$' and  `$\times$' denote locations of temperature and light sensors, respectively.}\vspace{-3mm}
	\label{fig:map}
	%\vspace{-5mm}
\end{figure}
\section{Signal-to-Noise Ratios for Jura Dataset}\label{datasets2}
\begin{table}[h]
	\centering
		\begin{tabular}{lccc}
			\hline
			& lg-Cd & Ni & lg-Zn\\
			\hline
			$\sigma^2_{s_i}$ & 2.2204 & 8.8280 & 2.3198 \\
			$\sigma^2_{n_i}$ & 0.0853 & 0.1130 & 0.0596 \\
			$\rho_i$  & \textbf{26.0305} & \textbf{78.1239} & 38.9228\\
			\hline
		\end{tabular}
	\caption{Signal-to-noise ratios $\rho_i$ of lg-Cd, Ni, and lg-Zn measurements for Jura dataset with $|U|=100$.}
	\label{tab:jura}
\end{table}

\fi
\end{document}